\documentclass[letterpaper, 10 pt, journal, twoside]{ieeetran}

\IEEEoverridecommandlockouts                              

\usepackage{graphics} 
\graphicspath{{figures/}}
\usepackage{epsfig}
\usepackage{times} 
\usepackage{siunitx}
\usepackage{amsmath} 
\usepackage{amssymb}  
\usepackage{hyperref}
\usepackage{subcaption}
\usepackage{booktabs}
\usepackage{multirow}
\usepackage[capitalise]{cleveref}
\Crefname{algorithm}{Alg.}{Algs.}
\Crefname{section}{Sec.}{Secs.}
\Crefname{equation}{Eq.}{Eqs.}
\Crefname{figure}{Fig.}{Figs.}
\usepackage{pgfplots}
\usepackage{pgfplotstable}
\usepgfplotslibrary{fillbetween}
\pgfplotsset{compat=1.17}
\usepackage{tikz}
\usepackage{comment}

\usetikzlibrary{positioning, arrows.meta, shapes, matrix}
\usepackage[normalem]{ulem}
\usepackage[
    sorting=none,
    giveninits=true,
    maxbibnames=6,
    maxcitenames=2,backend=biber
    ]{biblatex}
\newlength\myheight
\newcommand*\ccircled[1]{\settowidth{\myheight}{#1}%
    \raisebox{-.1\myheight}{\tikz[baseline=(char.base)]{%
        \node[shape=circle,draw,minimum size=\myheight*\myheight*.4,inner sep=1pt](char){#1};}}}

\title{Trajectory Prediction for Heterogeneous Agents: \\A Performance Analysis on Small \\and Imbalanced Datasets}

\author{Tiago Rodrigues de Almeida$^{1}$, Yufei Zhu$^{1}$, Andrey Rudenko$^{2}$,  \\ Tomasz P. Kucner$^{3}$, Johannes A. Stork$^{1}$, Martin Magnusson$^{1}$, Achim J. Lilienthal$^{1,4}$%
\thanks{Manuscript received: February, 13, 2024; Revised April, 27, 2024; Accepted May, 23, 2024. This paper was recommended for publication by Editor Gentiane Venture upon evaluation of the Associate Editor and Reviewers’ comments.
This work was supported by the Wallenberg AI, Autonomous Systems and Software Program (WASP) and by the EU Horizon 2020 No. 101017274 (DARKO). 
\textit{Tiago Rodrigues de Almeida and Yufei Zhu contributed equally to this work.} \textit{(Corresponding author: Tiago Rodrigues de Almeida.)}} 
\thanks{$^{1}$Tiago Rodrigues de Almeida, Yufei Zhu, Johannes A. Stork, Martin Magnusson, and Achim J. Lilienthal are with the Centre for Applied Autonomous Sensor Systems (AASS), \"Orebro University, Sweden 
        {\tt\small \{tiago.almeida, yufei.zhu, johannesandreas.stork, martin.magnusson, achim.lilienthal\}@oru.se}}
\thanks{$^{2}$ Andrey Rudenko is with the Robert Bosch GmbH, Corporate Research, Stuttgart, Germany
{\tt\footnotesize andrey.rudenko@de.bosch.com}}%
\thanks{$^{3}$Tomasz P. Kucner is with the School of Electrical Engineering Aalto University and with the Finnish Center for Artificial Intelligence (FCAI), Finland
{\tt\footnotesize tomasz.kucner@aalto.fi}}%
\thanks{$^{4}$Achim J. Lilienthal is also with the Technical University of Munich, Germany.
{\tt\footnotesize achim.j.lilienthal@tum.de}}%
\thanks{Digital Object Identifier (DOI): see top of this page.}
}

\newif\ifkeepremark

\keepremarkfalse

\usepackage{etoolbox}
\makeatletter
\def\nouppercase#1{{\let\uppercase\relax\let\MakeUppercase\relax #1}}

\def\ps@IEEEtitlepagestyle{%
  \def\@oddhead{%
    \hbox{}%
    \hfil
    \parbox[t]{.9\textwidth}{\centering\scriptsize
      \nouppercase{© 2024 IEEE. Personal use of this material is permitted. 
      Permission from IEEE must be obtained for all other uses, in any current or future media, 
      including reprinting/republishing this material for advertising or promotional purposes, 
      creating new collective works, for resale or redistribution to servers or lists, or reuse 
      of any copyrighted component of this work in other works. Please cite the paper as: T. R. de Almeida et al., "Trajectory Prediction for Heterogeneous Agents: A Performance Analysis on Small and Imbalanced Datasets," in IEEE Robotics and Automation Letters, 2024, doi: 10.1109/LRA.2024.3408510.}}%
    \hfil
    \thepage}%
  \let\@evenhead\@oddhead
  \let\@oddfoot\@empty
  \let\@evenfoot\@empty
}
\makeatother

\begin{document}

\maketitle

\begin{abstract}
    Robots and other intelligent systems navigating in complex dynamic environments should predict future actions and intentions of surrounding agents to reach their goals efficiently and avoid collisions. The dynamics of those agents strongly depends on their tasks, roles, or observable labels. Class-conditioned motion prediction is thus an appealing way to reduce forecast uncertainty and get more accurate predictions for heterogeneous agents. However, this is hardly explored in the prior art, especially for mobile robots and in limited data applications. In this paper, we analyse different class-conditioned trajectory prediction methods on two datasets. We propose a set of conditional pattern-based and efficient deep learning-based baselines, and evaluate their performance on robotics and outdoors datasets (TH\"{O}R-MAGNI and Stanford Drone Dataset). Our experiments show that all methods improve accuracy in most of the settings when considering class labels. More importantly, we  observe that there are significant differences when learning from imbalanced datasets, or in new environments where sufficient data is not available. In particular, we find that deep learning methods perform better on balanced datasets, but in applications with limited data, e.g., cold start of a robot in a new environment, or imbalanced classes, pattern-based methods may be preferable.
\end{abstract}

\begin{IEEEkeywords}
Human and Humanoid Motion Analysis and Synthesis, Human Detection and Tracking, Datasets for Human Motion, Deep Learning Methods
\end{IEEEkeywords}
\section{INTRODUCTION} \label{section-intro}

\IEEEPARstart{R}{eliable} and safe robot navigation in dynamic human-centered environments relies on anticipating the future 
behavior of other agents. In several domains, including autonomous driving (AD) and industrial mobile 
robotics, the motion planner must proactively consider the future positions 
of many heterogeneous agents to maintain safety standards~\cite{teja21}. In 
autonomous urban driving, 
various entities with distinct dynamic patterns, such as pedestrians, cyclists, and cars, navigate in a shared space.
In industrial environments, humans engaged in different tasks, such as transporting objects, interacting 
with robots, or walking in groups, may also present different motion patterns within the same spatial 
layout~\cite{almeida23}. The diversity of agents and their corresponding motion patterns pose significant 
challenges to current trajectory predictors~\cite{fang22}, leading to high prediction uncertainty, lower prediction accuracy, and thus to overly conservative motion planning~\cite{heuer2023proactive}. 

\begin{figure}[!t]
   \centering
   \includegraphics*[clip,trim=0mm 25mm 0mm 0mm, width=\linewidth]{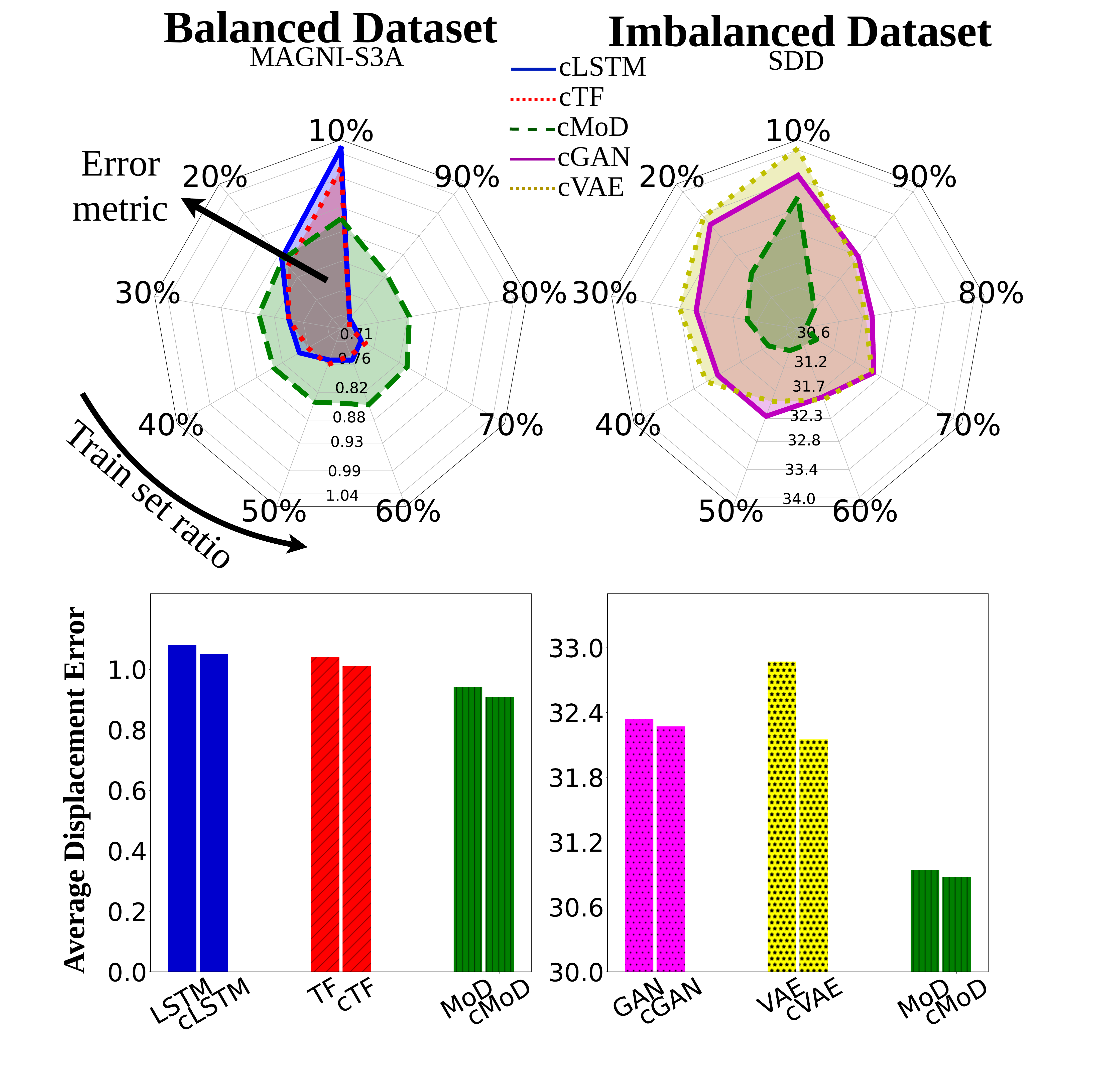}
   \vspace*{-7mm}
   \caption{Average Displacement Error of class-conditioned trajectory prediction methods across balanced and imbalanced datasets. 
   \textbf{Top:} In balanced datasets, at 
   low data regimes (10\% train set ratio), the pattern-based method (cMoD) is more accurate than deep learning methods (\textbf{left}). In imbalanced datasets, 
   cMoD is more accurate than deep generative models (\textbf{right}). Thus, the pattern-based method, cMoD, is more suitable 
   for low training data regimes and imbalanced datasets than its deep learning counterparts. \textbf{Bottom}: Class-conditioned 
   methods (c***) consistently outperform their unconditional counterparts across both datasets.}
   \label{fig:cover}
\vspace*{-6mm}
\end{figure}

In the context of mobile robots in dynamic environments, prior art has hardly explored class-aware prediction approaches, 
which is particularly evident given that human motion datasets with class or activity labels are still rare~\cite{schreiter2022magni}. Moreover, 
existing heterogeneous trajectory prediction methods tailored for AD do not transfer well to robotics settings, as they depend on domain-specific contextual features~\cite{chandra19}. 
Furthermore, robotics applications present unique challenges, such as the cold-start scenario, 
where a robot enters and continuously navigates a previously unseen environment with limited data~\cite{cui21}. Additionally, both robotics and AD domains may feature non-uniform class distributions, leading to decreased performance of deep learning-based trajectory prediction methods~\cite{almeida22}.
It is important to understand whether class-conditioned prediction methods can benefit in applications with scarce or imbalanced data, and if so, to what extent and under which specific circumstances. 


In this paper, we present an in-depth study of class-conditioned trajectory prediction methods
under different conditions. We extend a pattern-based approach CLiFF-LHMP~\cite{zhu2023clifflhmp}, which uses Maps of Dynamics~\cite{tomasz_survey23} (MoD), to introduce a class-conditioned variant, and similarly adapt several deep learning methods to include class labels. 
In contrast to previous methods~\cite{ma2019trafficpredict,adaptive_23_geng,heat_22_mo}, our proposed deep learning approaches are both memory and energy efficient as they do not require training or running individual modules per class.
We assess their performance across diverse training data conditions, considering both balanced and imbalanced datasets 
(where class proportions are uniform and non-uniform, respectively), and various amounts of training data. The study of 
imbalanced datasets is significant as deep learning methods may struggle to predict underrepresented classes, 
which is particularly impactful when these classes represent vulnerable road users such as pedestrians. The study of various training data 
amounts reflects a practical challenge in mobile robotics, where the system is deployed in new environments 
with limited acquired data yet requiring anticipation of other agents' movements for safe navigation.
We analyse heterogeneous agents prediction in two distinct datasets: the Stanford Drone Dataset (SDD)~\cite{robicquet16} with diverse road users outdoors, and the TH\"{O}R-MAGNI dataset~\cite{schreiter2022magni}, with mobile robots and human agents in a mockup indoor industrial 
environment. 
Through this comparative study, we aim to show the preferred methods for specific settings, quantifying their performance in 
different data 
regimes and class-imbalanced datasets.
\cref{fig:cover} outlines the main results of our study. 

In summary, we make the following contributions:

\begin{itemize}
    \item We establish a set of conditional Maps of Dynamics (MoD) and deep learning-based trajectory prediction baselines\footnote{Code available at \url{https://github.com/tmralmeida/class-cond-trajpred}} for outdoor mixed traffic scenarios (SDD) and an indoor mobile robot dataset (TH\"{O}R-MAGNI).
    \item We analyse the performance of four deep learning methods and an MoD approach that consider activity labels or agent classes in TH\"{O}R-MAGNI and SDD.
    \item 
    We show that class-conditioned methods outperform their unconditional counterparts in most cases. In addition, we show that MoD approaches are preferable over the deep generative methods for class-imbalanced datasets and superior to single-output deep learning methods in low data regimes.
\end{itemize}

\section{RELATED WORK} \label{section-relatedwork}

\subsection{Motion Prediction for Heterogeneous Agents}

The task of heterogeneous trajectory prediction involves estimating the future positions of an agent based on an observed trajectory, augmented by features describing the agent class, and optionally incorporating additional contextual factors, such as the obstacle maps. Deep learning has been widely applied to solve this problem~\cite{chandra19,unin_21_zheng,adaptive_23_geng}, in particular Graph Neural Networks 
in the context of Autonomous 
Driving~\cite{ma2019trafficpredict,salzmann20,Rainbow2021SemanticsSTGCNNAS,unin_21_zheng,heat_22_mo}. However, methods developed for predicting the motion of road agents do not transfer their assumptions when applied to other 
environments, such as intralogistic or public spaces. For instance, \emph{TraPhic}~\cite{chandra19} uses the shape of the road agent 
as a discriminative input feature for the various classes, which does not scale well to datasets with diverse human activities, such as TH\"{O}R~\cite{rudenko2020thor} or TH\"{O}R-MAGNI~\cite{schreiter2022magni}. Conversely, \emph{HAICU}~\cite{ivanovic22} uses the output of the perception 
module as a representation of the road agent's class, incorporating a continuous label distribution instead of a discrete value as the agent's type. 
Finally, \cite{asghar23} explores the use of dynamic Occupancy Grid Maps (OGMs) combined with semantic attributes to predict vehicle trajectories. However, it does not account for the heterogeneous entities typically present in road environments (e.g., pedestrians and cyclists). This limitation highlights the need for models that can be applied to different road users.

Alternative approaches involve individual deep learning modules for each agent 
class~\cite{ma2019trafficpredict,heat_22_mo,adaptive_23_geng} to account for the heterogeneity of the dataset. 
These methods require individual encoders and/or decoders for each class, which presents scalability challenges as the number of classes increases.
Conversely, in this paper we condition deep learning-based trajectory predictors on class embeddings, leading to a single model encompassing all classes.
\emph{Semantics-STGCNN}~\cite{Rainbow2021SemanticsSTGCNNAS} addresses imbalanced class proportions in 
heterogeneous motion trajectory datasets through a class-balancing loss function, yet this strategy 
struggles to achieve top performance in all classes.
To address issues of deep learning methods in imbalanced datasets, we propose a class-conditioned method based on Maps of Dynamics that is significantly less sensitive to class proportions than deep learning approaches and thus improves on its unconditional counterpart.

In this paper, we conduct an analysis of class-conditioned methods that are agnostic to the scene environment, allowing for applicability across different settings. 
Specifically, we evaluate four deep learning models and a Maps of Dynamics method~\cite{zhu2023clifflhmp}, along with their respective conditional 
counterparts, on two datasets of human motion (TH\"{O}R-MAGNI) and road agent trajectories (SDD). We argue these methods are well-suited for application in new environments to support safe mobile robot navigation.

\subsection{Heterogeneous Motion Trajectory Datasets}

Motion trajectory datasets reflect a variety of factors that describe the dynamics of the agents' movements. These factors commonly relate to (1) agent-agent interactions~\cite{salatiello21, kothari23}, 
providing insights into the social and interactive motion; 
(2) agent-environment interactions~\cite{kratzer_mogaze_2021}, describing specific environment-related events in trajectory data
or activities performed by the agent; (3) human-robot interaction (HRI)~\cite{karnan2022socially} 
supporting the development of social navigation methods. This paper focuses on trajectory prediction in
heterogeneous motion datasets containing various classes of agents. These classes include labeled agents such as cars, pedestrians, and bicyclists\cite{robicquet16}, 
or diverse human activities that influence the motion dynamics in a working environment~\cite{schreiter2022magni}.

Heterogeneous human motion datasets have been gathered across diverse environments, including road 
scenes~\cite{argoverse19,argoverse21}, university campuses~\cite{robicquet16,lcas_17,ehsan22}, surveilled 
outdoor areas~\cite{virat_11,ucla_15}, and indoor settings~\cite{rudenko2020thor,schreiter2022magni,kratzer_mogaze_2021}. 
In this paper, we tested our prediction methods on two datasets:
the well-established outdoor SDD~\cite{robicquet16} and the novel indoor TH\"{O}R-MAGNI dataset~\cite{schreiter2022magni}. We chose SDD and TH\"{O}R-MAGNI due to their distinct settings: imbalanced outdoor road agents and balanced human tasks in a robotics environment, respectively.

\section{METHODS} \label{section-methods}
\subsection{Problem Statement}
We frame the task of trajectory prediction as inferring a sequence of future states $\mathcal{T}$ with the input of an 
observation sequence $X$, and the class of the agent $c$. A state $s \in X$ of an agent is represented by the 2D Cartesian coordinates $(x,y)$ and the corresponding velocity vector $(v_x,v_y)$, i.e., $s = (x,y,v_x,v_y)$. Velocity can also be decomposed into 2D speed and orientation. The future sequence $\mathcal{T}$ can be composed of velocities $Y$ and positions $P$, depending on the method formulation.
After observing $O_p$ time steps, $T_p$ future states are predicted, 
i.e. $O_p = |X|$ and  $T_p = |\mathcal{T}|$.

\subsection{Deep Learning Models} \label{subsec:deeplearning}

In this section, we present the deep learning methods to predict motion trajectories considering agent classes. Our analysis includes both single-output trajectory predictors (one prediction per observed trajectory), namely Long Short-Term Memory (LSTM), specifically the RED method~\cite{becker18}, and Transformer-based, denoted by TF~\cite{almeida23}, as well as 
multiple-output approaches (multiple predictions per observed trajectory), including Generative Adversarial Networks (GANs) and Variational Autoencoders (VAEs). We use the single-output methods, RED and TF, and their respective class-conditioned counterparts cRED and cTF, as outlined in~\cite{almeida23}. The multiple-output approaches, GAN, VAE, and their respective conditioned counterparts cGAN and cVAE use Transformers-based encoders~\cite{vaswani17} (as described in this section). 

\subsubsection{Single-Output Trajectory Predictors} 
\begin{figure}[!t]
 \tikzset{myLine1/.style = {->, thick, >=stealth},
    mynarrownodes/.style = {node distance=0.75cm and 0.3cm},
    }
  \tikzstyle{bigbox} = [draw=blue!50, thick, fill=blue!20, rounded corners, rectangle]
  \tikzstyle{box} = [minimum size=0.1cm, rounded corners,rectangle split,rectangle split parts=2, inner sep=0.1ex, fill=gray!10]
  \tikzstyle{box3} = [minimum size=0.1cm, rounded corners,rectangle split,rectangle split parts=3, inner sep=0.1ex, fill=gray!10]
    \centering
      \begin{tikzpicture}
      \node[inner sep=0, anchor=center] (input2) at (-4,0) {$X$};
      \matrix[right=0.75cm of input2, row sep=0mm, column sep=2.5mm, inner sep=1mm, bigbox, every node/.style=box3](lstm2) {
            \node(embedding2){\footnotesize{\textbf{Embedding}} \vphantom{$\vcenter{\vspace{1.em}}$} \nodepart{second} \scriptsize{MLP}}; 
            &  \node(encoder2){\footnotesize{\textbf{Encoder}} \vphantom{$\vcenter{\vspace{1.em}}$} \nodepart{second} \scriptsize{LSTM} \nodepart{third} \scriptsize{Transformer}}; 
            & \node(decoder2){\footnotesize{\textbf{Decoder}} \vphantom{$\vcenter{\vspace{1.em}}$} \nodepart{second} \scriptsize{MLP}}; \\
        };
        \node[inner sep=0, anchor=center, above=0.34cm of embedding2] (role) {$c$};
        \node[inner sep=0, anchor=center, above=0.3cm of encoder2] (concat) {$\oplus$};
        \node[inner sep=0, anchor=center, right=0.75cm of lstm2] (output2) {$\hat{Y}$};

        \draw[myLine1] (embedding2.east) to[out=0, in=180] (encoder2.west);
        \draw[myLine1, dashed] (encoder2.north) to[out=90, in=-90] (concat.south);
        \draw[myLine1, dashed] (concat.east) to[out=0, in=90] (decoder2.north);
        \draw[myLine1, dashed] (role.east) to[out=0, in=180] (concat.west);

        \draw[myLine1] (input2.east) to[out=0, in=180] (lstm2.west);
        \draw[myLine1] (lstm2.east) to[out=0, in=180] (output2.west);
      
      \end{tikzpicture}
\caption{Single-output unconditional and conditional methods (dashed lines).}
\vspace{-4mm}
\label{fig:sing_out_overview}
\end{figure}
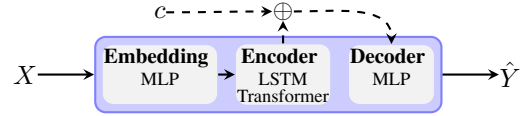

The first step is embedding $X$ using a 
 Multi-Layer Perceptron (MLP) network (see \cref{fig:sing_out_overview}). Additionally, cRED and cTF embed the integer class label $c$ with an embedding layer. Subsequently, for RED and cRED, the embedded input vector 
 passes through an LSTM layer, while for TF and cTF, the encoded input vector undergoes a Transformer-based encoder. An MLP-based decoder then generates the predicted sequence of velocity vectors from the encoded vectors. For conditional variants (cRED and cTF), class embeddings are concatenated with the temporal features before decoding. 
 We train single-output networks with the Mean Squared Error (MSE) loss:
\begin{equation}
      L_{T}(P, \hat{P}) = \frac{1}{T_{p}} \sum_j^{T_p} \lVert p^j -\hat{p}^j  \rVert_2,
\label{eq:l_traj}
\end{equation}
where $\hat{P}$ represents the estimated sequence of positions, $p^{j}$ the ground truth position at time step $j$ and $\hat{p}^j$ the corresponding predicted position.

\subsubsection{GAN-based Trajectory Predictors}
A GAN aims to reconstruct 
the generative process of the underlying input data using two modules: the generator ($G$) and the discriminator ($D$). The generator maps the input $X$ and a latent random vector $\mathbf{z_G}$ to a realistic future set of velocities $Y$. 
We sample the latent vector from a standard normal Gaussian distribution. 
Simultaneously, the discriminator differentiates both real and generated future velocity vectors, $Y$ and 
$\hat{Y}$, respectively. This adversarial 
training scenario is essential for producing multiple plausible trajectories. 
In cGAN, both the generator and discriminator incorporate the trajectory class as an additional input.
We optimize the GAN and cGAN discriminators using the binary cross-entropy loss, while the GAN generator is optimized with 
a weighted sum given by:
\begin{equation}
      L_{G} 
      = 
      \lambda_1 L_{T} + \lambda_2 
      \left(\frac{1}{2} \mathbb{E}[(D(Y) - 1)^2] + \frac{1}{2} \mathbb{E}[D(\hat{Y})^2]\right),
\end{equation}
where $\lambda_1$ and $\lambda_2$  are the weights applied to the MSE term $L_{T}$ (Eq.~\ref{eq:l_traj}) and 
to the GAN loss, respectively. 
For the conditional variant (cGAN), the class is additionally fed as input to both the generator and the discriminator.

\begin{figure}[t]
  \centering
  \tikzset{myLine1/.style = {->, thick, >=stealth},
    mynarrownodes/.style = {node distance=0.75cm and 0.3cm},
    }
  \tikzstyle{gen_dashed} = [minimum width=2.32cm, minimum height=3cm, draw=black, dashed, rectangle]
  \tikzstyle{disc_dashed} = [minimum width=1.55cm, minimum height=3cm, draw=black, dashed, rectangle]
  \tikzstyle{bigbox} = [draw=black!50, thick, fill=gray!50, rounded corners, rectangle]
  \tikzstyle{box} = [minimum size=0.2cm, rounded corners,rectangle split,rectangle split parts=2, inner sep=0.1ex, fill=gray!10]
  \tikzstyle{box_decoder} = [minimum size=0.2cm, rounded corners,rectangle split,rectangle split parts=4, inner sep=0.1ex, fill=gray!10]
  \begin{subfigure}{0.48\columnwidth}
    \centering
      \begin{tikzpicture}
      \node[inner sep=0, anchor=center] (input) at (-4,2) {$X$};
      \node[style=gen_dashed, fill=yellow!5,minimum width=2.1cm](gen_dash) at (-4.27,0.1){}; 
      \node[inner sep=0, anchor=center] (g_str) at (-5.1,1.4) {$G$};
      \node[right=0.06cm of gen_dash, style=disc_dashed, fill=blue!5](disc_dash){}; 
      \node[inner sep=0, anchor=center] (d_str) at (-2.95,1.4) {$D$};
      
      \matrix[below=0.5cm of input, row sep=2.5mm, column sep=0mm, inner sep=1mm, bigbox](gen) {
            \node(embedding)[style=box]{\footnotesize{\textbf{Embed.}} \vphantom{$\vcenter{\vspace{1.em}}$} \nodepart{second} \scriptsize{MLP}}; 
            \\  \node(encoder)[style=box]{\footnotesize{\textbf{Encoder}} \vphantom{$\vcenter{\vspace{1.em}}$} \nodepart{second} \scriptsize{Transformer}}; 
            \\ \node(decoder)[style=box]{\footnotesize{\textbf{Decoder}} \vphantom{$\vcenter{\vspace{1.em}}$} \nodepart{second} \scriptsize{MLP}}; \\
        };
        \node[inner sep=0, anchor=center, below=0.6cm of gen] (output) {$\hat{Y}$};
        \node[inner sep=0, anchor=center, left=0.34cm of embedding] (noise) {$z_G$};
        \node[inner sep=0, anchor=center, left=0.3cm of encoder] (concat) {$\oplus$};

        \draw[myLine1] (embedding.south) to[out=-90, in=90] (encoder.north);
        \draw[myLine1] (encoder.west) to[out=180, in=0] (concat.east);
        \draw[myLine1] (noise.south) to[out=-90, in=90] (concat.north);
        \draw[myLine1] (concat.south) to[out=-90, in=180] (decoder.west);

        \node[inner sep=0, anchor=center, right=0.9cm of input] (inputd) {$Y$ or $\hat{Y}$};
        \matrix[below=1.25cm of inputd, row sep=2.5mm, column sep=0mm, inner sep=1mm, bigbox](disc) { 
          \node(class)[style=box_decoder]{\footnotesize{\textbf{Classifier}} \vphantom{$\vcenter{\vspace{1.em}}$}\nodepart{second} \scriptsize{Transformer} \nodepart{third} \scriptsize{+} \nodepart{fourth} \scriptsize{MLP}}; \\
        };
        \node[inner sep=0, anchor=center, below=1.2cm of disc] (score) {$s$};

        \draw[myLine1] (input.south) to[out=-90, in=90] (gen.north);
        \draw[myLine1] (gen.south) to[out=-90, in=90] (output.north);
        \draw[myLine1] (inputd.south) to[out=-90, in=90] (disc.north);
        \draw[myLine1] (disc.south) to[out=-90, in=90] (score.north);
      
      \end{tikzpicture}
  \end{subfigure}
  \hfill
  \begin{subfigure}{0.48\columnwidth}
    \centering
      \begin{tikzpicture}
      \node[inner sep=0, anchor=center] (input) at (-3.7,2) {$\ccircled{$c$}\oplus X$};
      \node[style=gen_dashed, fill=yellow!15](gen_dash) at (-4.1,0.1){}; 
      \node[inner sep=0, anchor=center] (g_str) at (-4.96,1.4) {$cG$};
      \node[right=0.04cm of gen_dash, style=disc_dashed, fill=blue!15](disc_dash){}; 
      \node[inner sep=0, anchor=center] (d_str) at (-2.5,1.4) {$cD$};

      \matrix[below=0.5cm of input, row sep=2.5mm, column sep=0mm, inner sep=1mm, bigbox](gen) {
            \node(embedding)[style=box]{\footnotesize{\textbf{Embed.}} \vphantom{$\vcenter{\vspace{1.em}}$} \nodepart{second} \scriptsize{MLP}}; 
            \\  \node(encoder)[style=box]{\footnotesize{\textbf{Encoder}} \vphantom{$\vcenter{\vspace{1.em}}$} \nodepart{second} \scriptsize{Transformer}}; 
            \\ \node(decoder)[style=box]{\footnotesize{\textbf{Decoder}} \vphantom{$\vcenter{\vspace{1.em}}$} 
            \nodepart{second} \scriptsize{MLP}}; \\
        };
        \node[inner sep=0, anchor=center, below=0.6cm of gen] (output) {$\hat{Y}$};
        \node[inner sep=0, anchor=center, left=0.34cm of embedding] (noise) {$z_G$};
        \node[inner sep=0, anchor=center, left=0.3cm of encoder] (concat) {$\oplus$};
        
        \draw[myLine1] (embedding.south) to[out=-90, in=90] (encoder.north);
        \draw[myLine1] (encoder.west) to[out=180, in=0] (concat.east);
        \draw[myLine1] (noise.south) to[out=-90, in=90] (concat.north);
        \draw[myLine1] (concat.south) to[out=-90, in=180] (decoder.west);

        \node[inner sep=0, anchor=center, right=0.95cm of input] (inputd) {$c$};
        \node[inner sep=0, anchor=center, above=0.15mm of inputd] (cat) {$\oplus$};
        \node[inner sep=0, anchor=center, above=0.1mm of cat] (ys) {$Y$ or $\hat{Y}$};
        \matrix[below=1.25cm of inputd, row sep=2.5mm, column sep=0mm, inner sep=1mm, bigbox](disc) { 
          \node(class)[style=box_decoder]{\footnotesize{\textbf{Classifier}} \vphantom{$\vcenter{\vspace{1.em}}$}\nodepart{second} \scriptsize{Transformer} \nodepart{third} \scriptsize{+} \nodepart{fourth} \scriptsize{MLP}}; \\
        };
        \node[inner sep=0, anchor=center, below=1.2cm of disc] (score) {$s$};
        \draw[myLine1] (input.west) -| (-5.2, 0.5) |- (concat.west);

        \draw[myLine1] (input.south) to[out=-90, in=90] (gen.north);
        \draw[myLine1] (gen.south) to[out=-90, in=90] (output.north);
        \draw[myLine1] (inputd.south) to[out=-90, in=90] (disc.north);
        \draw[myLine1] (disc.south) to[out=-90, in=90] (score.north);
      
      \end{tikzpicture}
  \end{subfigure}
  \caption{GAN-based models: unconditional GAN (left) and cGAN (right).}
  \label{fig:gan_based}
  \vspace{-17pt}
\end{figure}
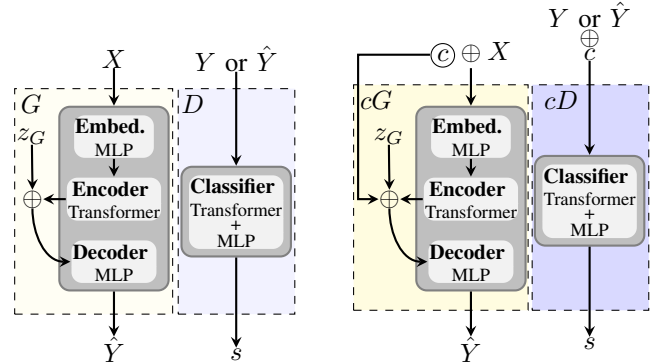

\cref{fig:gan_based} illustrates the network configurations for GAN and cGAN models. In general, the generators have the same layer configuration as the TF model. The difference is the latent vector, $\mathbf{z_G}$, which is concatenated with the temporal features from the transformer and passed to the decoder. Analogous to~\cite{kothari23}, the discriminator comprises a 
transformer-encoder network and a MLP in the last layer. For cGAN, both the generator and the discriminator concatenate $X$ to the agent's class embedding. The generator also concatenates the class embeddings to the input of the decoder.

\subsubsection{VAE-based Trajectory Predictors}

Generative VAE-based predictors consist of two main networks: the prior network $p_\theta$ and the recognition network $q_\phi$. The prior network maps the input state $X$ and a latent vector $\mathbf{z_V}$ to the predicted trajectory $Y$, while the recognition network learns to map the ground truth trajectory $Y$ to the parameters of a Gaussian distribution, representing a lower-dimensional latent space. We adopt 
a standard normal Gaussian as the prior for the distribution of future trajectories. The Kullback-Leibler (KL) divergence is used to align the learned distribution to the prior, contributing to the VAE's loss function:
\begin{equation}
      L_{V} = \beta_1 ~ L_T 
      - \beta_2 ~ D_\mathrm{KL}[q_{\phi}(z_V | Y) \rVert p_{\theta}(z_V | X)], 
  \end{equation}
  where $\beta_1$ and $\beta_2$ are the weights applied to the MSE and KL terms, respectively. For the conditional variant (cVAE), the agent's class is added as input to both $p_\theta$ and $q_\phi$.

\begin{figure}[t]
\centering
\tikzset{myLine1/.style = {->, thick, >=stealth},
  mynarrownodes/.style = {node distance=0.75cm and 0.3cm},
  }
\tikzstyle{q_dashed} = [minimum width=1.65cm, minimum height=3cm, draw=black, dashed, rectangle]
\tikzstyle{bigbox} = [draw=black!50, thick, fill=gray!50, rounded corners, rectangle]
\tikzstyle{box} = [minimum size=0.2cm, rounded corners,rectangle split,rectangle split parts=2, inner sep=0.1ex, fill=gray!10]
\tikzstyle{box_dec} = [minimum size=0.2cm, rounded corners,rectangle split,rectangle split parts=2, inner sep=0.1ex, fill=gray!30, draw=gray!30]
\begin{subfigure}{0.48\columnwidth}
  \centering
    \begin{tikzpicture}
    \node[inner sep=0, anchor=center] (input) at (-6,2) {$X$};
    \node[style=q_dashed, fill=white](q_dash) at (-4.0,0.25){}; 
    \node[inner sep=0, anchor=center] (q_str) at (-3.4,1.6) {$q_{\phi}$};
    \node[inner sep=0, anchor=center] (q_str) at (-5.7,1.6) {$p_\theta$};

    \matrix[below=0.5cm of input, row sep=2.5mm, column sep=0mm, inner sep=1mm, bigbox](vae) {
          \node(embedding1)[style=box]{\footnotesize{\textbf{Embed.}} \vphantom{$\vcenter{\vspace{1.em}}$} \nodepart{second} \scriptsize{MLP}}; 
          \\  \node(encoder)[style=box]{\footnotesize{\textbf{Encoder}} \vphantom{$\vcenter{\vspace{1.em}}$} \nodepart{second} \scriptsize{Transformer}}; \\
      };
      \node(decoder)[below=0.3cm of vae,style=box_dec]{\footnotesize{\textbf{Decoder}}\nodepart{second} \scriptsize{MLP}};

      \node[inner sep=0, anchor=center, below=0.5cm of decoder] (output) {$\hat{Y}$};
      
      \draw[myLine1] (embedding1.south) to[out=-90, in=90] (encoder.north);
      \draw[myLine1] (input.south) to[out=-90, in=90] (vae.north);
      \draw[myLine1] (decoder.south) to[out=-90, in=90] (output.north);

      \node[inner sep=0, anchor=center, right=1.7cm of input] (inputr) {$Y$};
      \matrix[below=0.5cm of inputr, row sep=2.5mm, column sep=0mm, inner sep=1mm, bigbox](recog) {
          \node(embedding_r)[style=box]{\footnotesize{\textbf{Embed.}} \vphantom{$\vcenter{\vspace{1.em}}$} \nodepart{second} \scriptsize{MLP}}; 
          \\  \node(encoder_r)[style=box]{\footnotesize{\textbf{Encoder}} \vphantom{$\vcenter{\vspace{1.em}}$} \nodepart{second} \scriptsize{Transformer}}; 
          \\ \node(out_r)[style=box]{\footnotesize{\textbf{Output}} \vphantom{$\vcenter{\vspace{1.em}}$} 
          \nodepart{second} \scriptsize{Linear}}; \\
      };

      \node[inner sep=0, anchor=center, below=0.6cm of out_r] (noise) {$\hat{z_V}$};
      \node[inner sep=0, anchor=center, right=0.28cm of decoder] (concat) {$\oplus$};
      \draw[myLine1] (embedding_r.south) to[out=-90, in=90] (encoder_r.north);
      \draw[myLine1] (inputr.south) to[out=-90, in=90] (recog.north);
      \draw[myLine1] (recog.south) to[out=-90, in=90] (noise.north);
      \draw[myLine1] (noise.west) to[out=-180, in=-90] (concat.south);
      \draw[myLine1] (encoder.east) to[out=0, in=90] (concat.north);
      \draw[myLine1] (concat.west) to[out=-180, in=0] (decoder.east);
      \draw[myLine1] (encoder_r.south) to[out=270, in=90] (out_r.north);

    \end{tikzpicture}
\end{subfigure}
\hfill
  \begin{subfigure}{0.48\columnwidth}
  \centering
    \begin{tikzpicture}
    \node[inner sep=0, anchor=center] (input) at (-6,2) {$X\oplus$ \ccircled{$c$}};
    \node[style=q_dashed, fill=white](q_dash) at (-3.82,0.25){}; 
    \node[inner sep=0, anchor=center] (q_str) at (-3.3,1.6) {$q_{\phi}$};
    \node[inner sep=0, anchor=center] (q_str) at (-5.7,1.6) {$p_\theta$};

    \matrix[below=0.5cm of input, row sep=2.5mm, column sep=0mm, inner sep=1mm, bigbox](vae) {
          \node(embedding1)[style=box]{\footnotesize{\textbf{Embed.}} \vphantom{$\vcenter{\vspace{1.em}}$} \nodepart{second} \scriptsize{MLP}}; 
          \\  \node(encoder)[style=box]{\footnotesize{\textbf{Encoder}} \vphantom{$\vcenter{\vspace{1.em}}$} \nodepart{second} \scriptsize{Transformer}}; \\
      };
      \node(decoder)[below=0.3cm of vae,style=box_dec]{\footnotesize{\textbf{Decoder}}\nodepart{second} \scriptsize{MLP}};
      
      \node[inner sep=0, anchor=center, below=0.5cm of decoder] (output) {$\hat{Y}$};
      
      \draw[myLine1] (embedding1.south) to[out=-90, in=90] (encoder.north);
      \draw[myLine1] (input.south) to[out=-90, in=90] (vae.north);
      \draw[myLine1] (decoder.south) to[out=-90, in=90] (output.north);

      \node[inner sep=0, anchor=center, right=1.2cm of input] (inputr) {$Y\oplus c$};
      \matrix[below=0.5cm of inputr, row sep=2.5mm, column sep=0mm, inner sep=1mm, bigbox](recog) {
          \node(embedding_r)[style=box]{\footnotesize{\textbf{Embed.}} \vphantom{$\vcenter{\vspace{1.em}}$} \nodepart{second} \scriptsize{MLP}}; 
          \\  \node(encoder_r)[style=box]{\footnotesize{\textbf{Encoder}} \vphantom{$\vcenter{\vspace{1.em}}$} \nodepart{second} \scriptsize{Transformer}}; 
          \\ \node(out_r)[style=box]{\footnotesize{\textbf{Output}} \vphantom{$\vcenter{\vspace{1.em}}$} 
          \nodepart{second} \scriptsize{Linear}}; \\
      };

      \node[inner sep=0, anchor=center, below=0.6cm of out_r] (noise) {$\hat{z_V}$};
      \node[inner sep=0, anchor=center, right=0.21cm of decoder] (concat) {$\oplus$};
      \draw[myLine1] (input.east) -| (-4.78, 0) |- (concat.east);
      \draw[myLine1] (embedding_r.south) to[out=-90, in=90] (encoder_r.north);
      \draw[myLine1] (inputr.south) to[out=-90, in=90] (recog.north);
      \draw[myLine1] (recog.south) to[out=-90, in=90] (noise.north);
      \draw[myLine1] (noise.west) to[out=-180, in=-90] (concat.south);
      \draw[myLine1] (encoder.east) to[out=0, in=90] (concat.north);
      \draw[myLine1] (concat.west) to[out=-180, in=0] (decoder.east);
      \draw[myLine1] (encoder_r.south) to[out=270, in=90] (out_r.north);

    \end{tikzpicture}
\end{subfigure}
\caption{VAE-based models: unconditional VAE (left) and cVAE (right). The recognition network ($q_{\phi}$, enclosed with dashed border) is 
solely available during training.}
\label{fig:vae_based}
\vspace*{-4mm}
\end{figure}
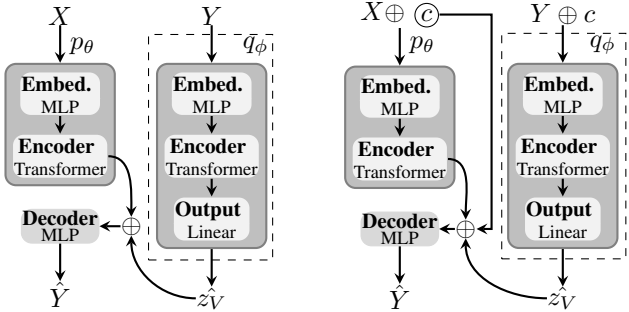

\cref{fig:vae_based} shows the network configurations for the VAE and cVAE models. The predictor's network configuration is identical to the generator in the GAN and cGAN models. The difference lies in the training process, where the latent vector $\mathbf{z_V}$ is sampled based on parameters generated by the recognition network ($q_\phi$). The recognition network processes the ground truth prediction akin to $p_\theta$ but concludes with two linear layers producing the Gaussian parameters.

\subsection{Pattern-based Trajectory Predictors}
Maps of Dynamics encode spatial or spatio-temporal motion patterns as a feature of the environment~\cite{kucner2017enabling, tomasz_survey23}. By generalizing velocity observations, human dynamics can be represented through flow models. Prior work proposes CLiFF-LHMP~\cite{zhu2023clifflhmp}, which exploits MoDs for long-term human motion prediction. It uses a multi-modal probabilistic representation of a velocity field (CLiFF-map), which is built from observations of human motion, and employs Semi-Wrapped Gaussian mixture models (SWGMM) to capture local velocity distributions. 
This method implicitly accounts for obstacle layouts and predicts trajectories that follow the environment's complex topology. CLiFF-LHMP excels in predicting up to \SI{50}{\second} ahead, \cite{zhu2023clifflhmp}, even with sparse, incomplete, and very limited training data \cite{zhu2023data}.

\begin{figure}
\centering
\includegraphics[clip,trim=0mm 0mm 0mm 0mm,height=29mm]{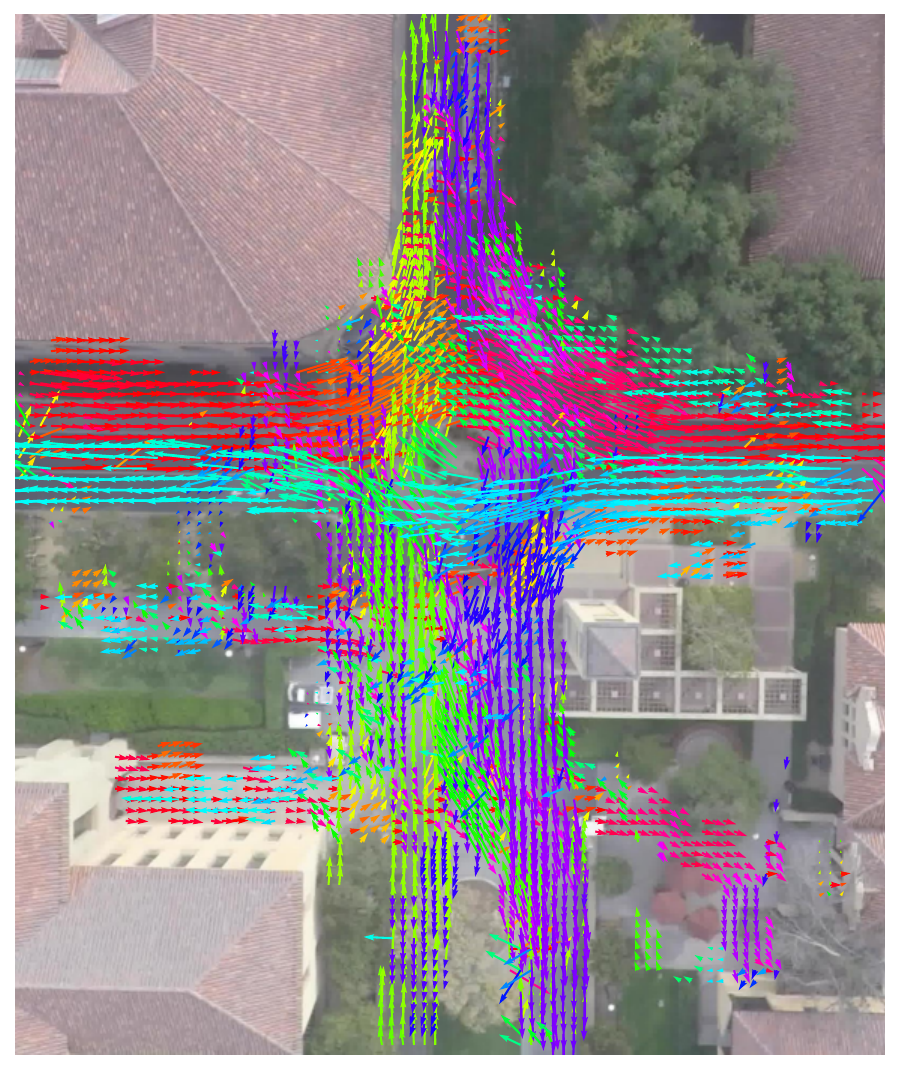}
\includegraphics[clip,trim=0mm 0mm 0mm 0mm,height=29mm]{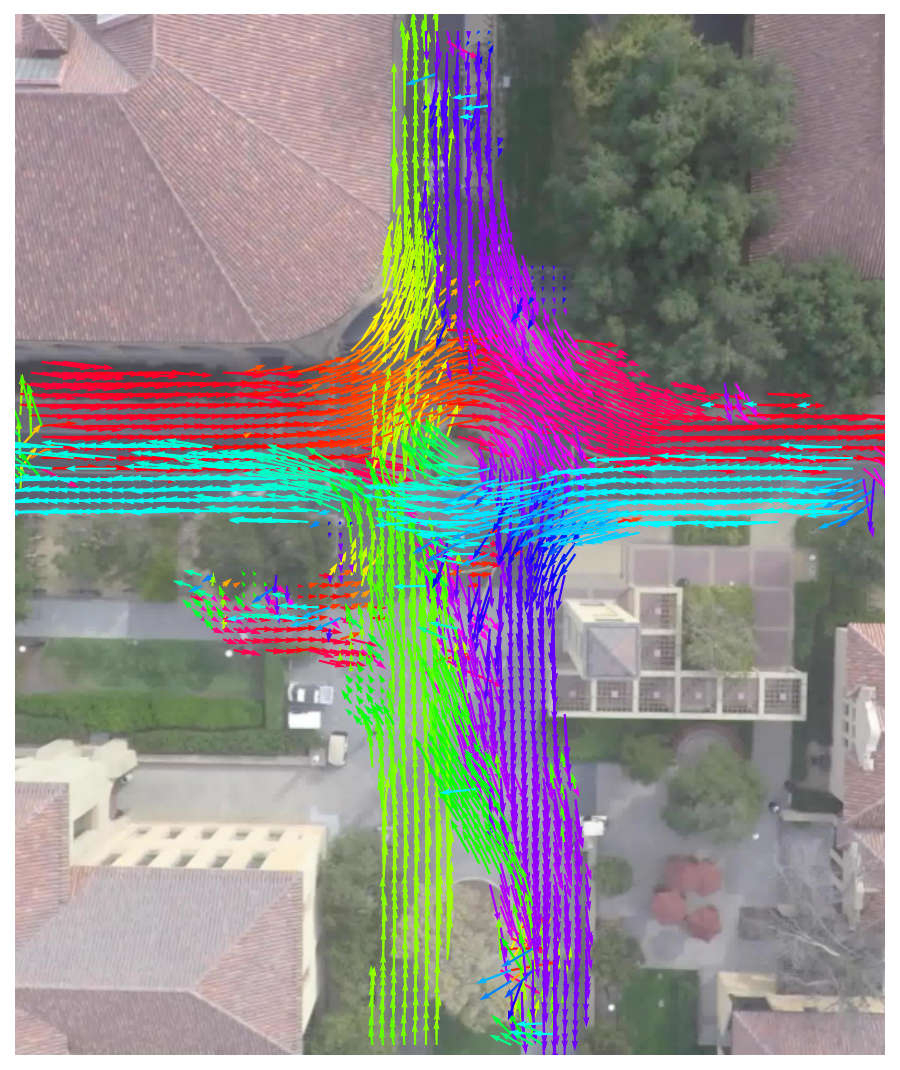}
\includegraphics[clip,trim=0mm 0mm 0mm 0mm,height=29mm]{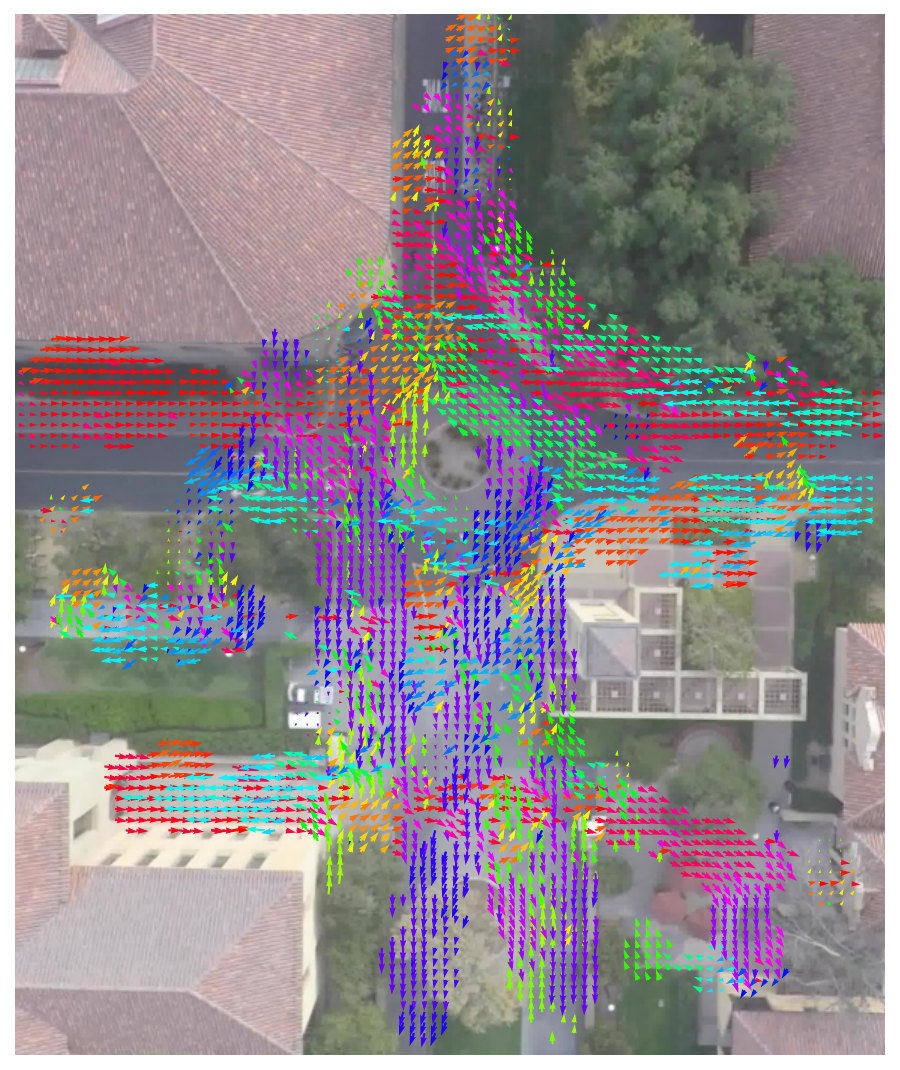}%
\includegraphics[clip,trim=195mm 15mm 2mm 12mm,height=29mm]{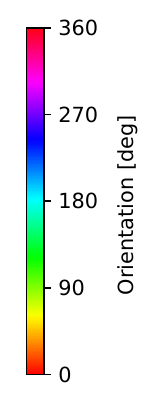}
\vspace{-5pt}
\caption{
General and class-conditioned CLiFF-maps in the {\em DeathCircle} scene of the SDD dataset. \textbf{Left:} all classes combined,  \textbf{middle:}: {\em Bicyclist} class, \textbf{right:} {\em Pedestrian} class. Colored arrows depict the mean speed (length) and direction (orientation) within the SWGMM of CLiFF-map, highlighting distinct motion patterns for different classes.
}
\label{fig:CLiFF-mapSDD}
\vspace*{-5mm}
\end{figure}

\begin{figure}[!t]
    \centering
    \includegraphics*[width=0.9\linewidth]{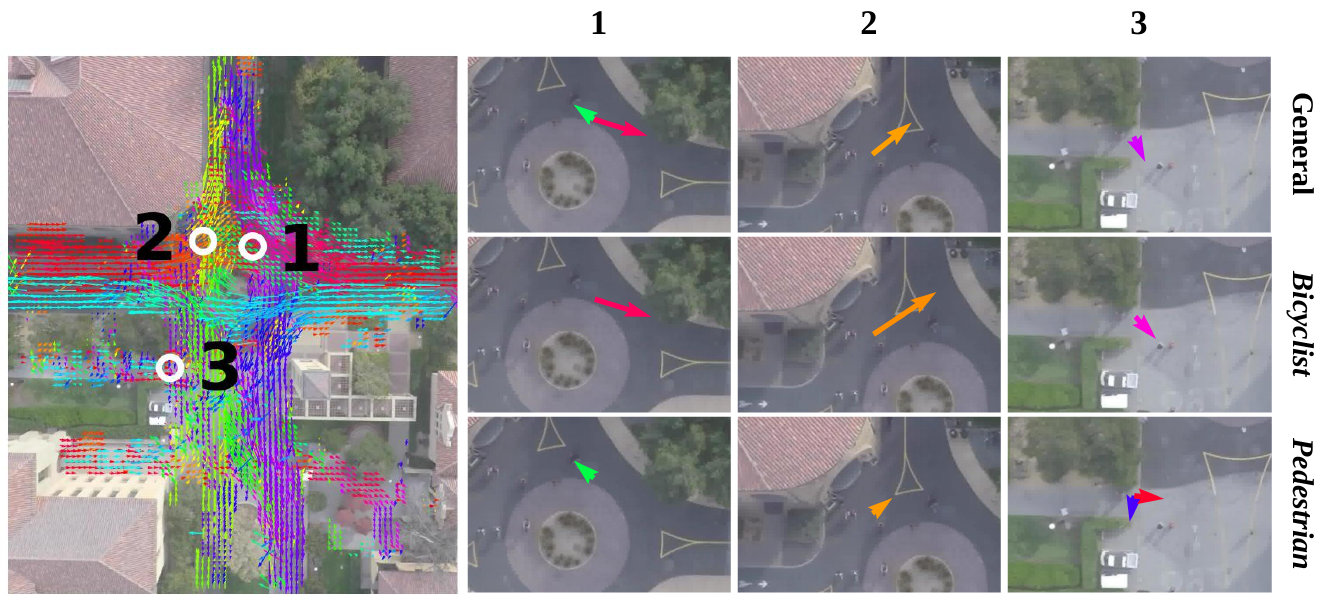}
    \vspace*{-1mm}
    \caption{CLiFF-maps at example locations in SDD ~\cite{robicquet16}. Both general and class-conditioned CLiFF-maps of {\em Bicyclist} and {\em Pedestrian} of three locations are shown on the \textbf{right}. General CLiFF maps may depict combinations of multiple classes (point 1) or median speed and orientation (points 2 and 3).}
\label{fig:point-example-sdd}
\vspace*{-5mm}
\end{figure}

In \cite{zhu2023clifflhmp}, a single CLiFF-map is used for all predicted trajectories, irrespective of the agent class. However, their motion patterns often differ, as shown in Fig.~\ref{fig:CLiFF-mapSDD} 
and further detailed in Fig.~\ref{fig:point-example-sdd}.
To address this, we introduce a class-conditioned CLiFF-map that differentiates the motion patterns representation to specific agent classes.

In a class-conditioned CLiFF-map, individual CLiFF-maps $\Xi_c$ are built for each agent class using their specific trajectories. For agent class $c$, we estimate $\mathcal{T}$ by sampling a velocity from $\Xi_c$ within the sampling radius $r_s$ for each prediction time step $t$. This velocity is then refined using a biased version of the Constant Velocity Model (CVM), following the same estimation process as the original CLiFF-LHMP, which is briefly described in the following. We refer the reader to \cite{zhu2023clifflhmp} for more details. The velocity prediction at time step $t$ is updated by biasing the last time step velocity with the sampled one as $ \rho_t = \rho_{t-1} + (\rho_s - \rho_{t-1}) \cdot Kn(\rho_{s} - \rho_{t-1}), 
\theta_t = \theta_{t-1} + (\theta_s - \theta_{t-1}) \cdot Kn(\theta_{s} - \theta_{t-1})$, where $\rho$ and $\theta$ represent speed and heading orientation of the agent, respectively. The kernel function $Kn$, defined as $Kn(x) =  e ^ {-\beta \left\Vert x \right\Vert ^ 2}$, modulates the influence of the sampled velocity. Using kernel $Kn$, the MoD term is scaled by the deviation between sampled and current velocities according to the CVM. The MoD is trusted less if it deviates more from the current velocity. Parameter $\beta$ controls the reliance on the MoD versus the CVM, with a lower $\beta$ favoring the velocity sampled from the MoD.


\section{EXPERIMENTS} \label{section-experiments}

\subsection{Datasets}
In this study, we evaluate and compare the performance of the trajectory prediction methods described in \cref{section-methods} on two datasets, TH\"{O}R-MAGNI \cite{schreiter2022magni} and SDD \cite{robicquet16}. 
Importantly for our analysis, there is a substantial difference in class proportions between the two datasets. 
Specifically, SDD shows a noticeable class imbalance compared to TH\"{O}R-MAGNI. This inter-dataset class imbalance poses a significant challenge to accurate trajectory prediction. We analyze how this challenge is handled by the two categories of predictors: deep learning models and MoDs approaches.

\textbf{TH\"{O}R-MAGNI} 
includes 3.5 hours of human motion data in a laboratory setting with static and mobile robots. In some scenarios, people are assigned tasks such as moving objects (boxes, buckets, poster stands) which significantly influence their motion patterns, especially the velocity profiles~\cite{almeida23}. This paper focuses on Scenarios 2, 3A, and 3B, which include 30 participants in 1.5 hours of motion. Five distinct agent roles are recorded in these scenarios: {\em Carrier--Large Object}, {\em Visitors--Group}, {\em Visitors--Alone}, {\em Carrier--Bucket}, and {\em Carrier--Box}, with corresponding sample proportions of 25.7\%, 23.6\%, 22.7\%, 14.1\%, and 13.9\%.

\textbf{SDD} 
encompasses 5 hours of heterogeneous trajectory data from 60 videos recorded on the Stanford University campus. 
It includes trajectories of bicyclists, pedestrians, skateboarders, carts, cars, and buses. 
Notably, certain classes such as {\em Bicyclist} and {\em Pedestrian} coexist in shared spaces but exhibit distinct movement patterns (e.g. bicyclists typically move faster). The dataset provides agent coordinates in pixel values. For our evaluation, we choose videos that contain at least two classes of agents and have above 10 trajectories per class, resulting in 7 scenes 
with cumulatively 3 agent classes: {\em Pedestrian}, {\em Bicyclist} and {\em Car} with corresponding sample proportions of 64.6\%, 34.3\%, and 1.1\%.

\subsection{Implementation Details}
To evaluate the predictors, we employed a repeated random sub-sampling validation method. 
For each iteration, we randomly selected $p\%$ of the dataset for training and used the remaining $(100-p)\%$ for testing. This process was repeated ten times, with the selection of test and training data being independently randomized in each iteration. 
In the accuracy analysis (\cref{sec:acc}), we set $p=90$. In the data efficiency analysis (\cref{sec:data_eff}), we decreased the percentage of data used for training from $p=90$ to $p=10$ in steps of $10$. Following current trajectory prediction benchmarks~\cite{Kothari2020HumanTF}, 
we set $O_p = 8$ and $T_p = 12$.

\textbf{For deep learning-based predictors}: 
We maintained a uniform hyperparameter setting to ensure a fair 
comparison. The training process for all networks extended to a maximum of $100$ epochs with early stopping after $20$ epochs with no improvement. We optimize the networks with the Adam optimizer~\cite{adam_15_optim}, a learning rate of $1\mathrm{e}{-3}$, and a batch size of $32$. We also reduce the learning rate on the plateau of the validation loss during training (patience set to $5$ epochs). 

\begin{figure}[t]
    \centering
    \includegraphics[width=\linewidth]{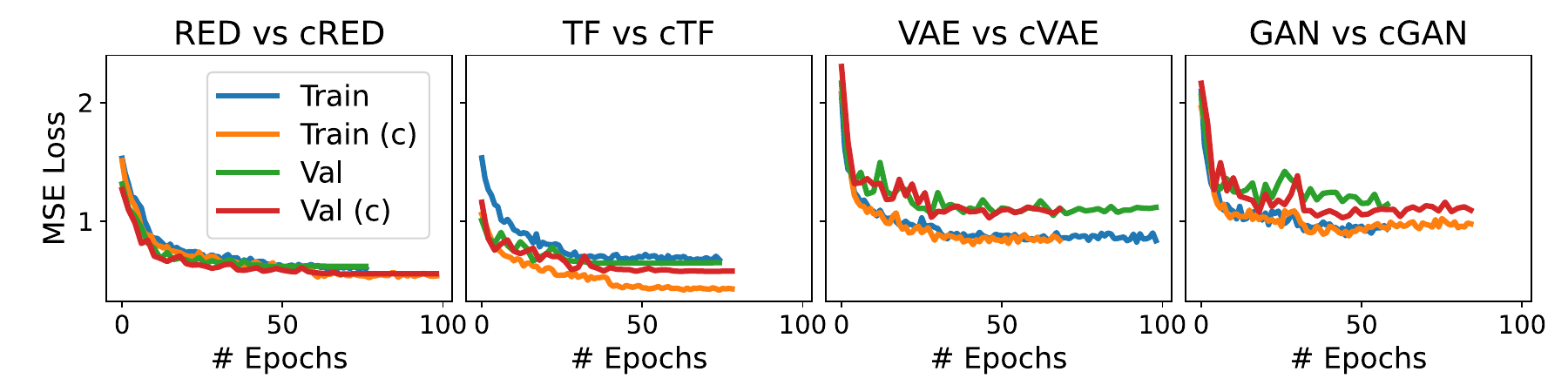}
    \caption{MSE training curves examples for the deep learning models, where (c) denotes conditional variants and \emph{Val} the validation curve.}
    \label{fig:mse_curves}
    \vspace{-15pt}
\end{figure}

For training generative models, including GAN, cGAN, VAE, and cVAE, we have standardized the weights in their 
respective loss functions. Consequently, $\lambda_1 = \beta_1 = 2$ and $\lambda_2 = \beta_2 = 1$, indicating a 
preference for the reconstruction of predictions based on the MSE term in the loss functions.

These hyperparameters and the networks' configurations described in \cref{subsec:deeplearning} allow training without overfitting, as shown by the loss curve examples in \cref{fig:mse_curves}.
Finally, each model receives as input state the position concatenated with the velocity vector for TH\"{O}R-MAGNI scenarios. In contrast, for SDD, the velocity vector alone is used as input due to the aggregation of diverse scenes, making the position an irrelevant input feature.

\textbf{For MoD-based predictors}:
Identical parameters are used for both the class-conditioned and the general CLiFF-LHMP. The CLiFF-map grid resolutions for the SDD dataset and the TH\"{O}R-MAGNI dataset are 20 pixels and \SI{0.2}{\metre}, respectively. The sampling radius $r_s$ is adjusted for each dataset to match the CLiFF map grid resolution. The kernel parameter is set to 5 for all experiments. In the figures and tables presenting the results, CLiFF-LHMP is denoted as MoD and class-conditioned CLiFF-LHMP is denoted as cMoD.

\subsection{Evaluation Metrics}
To compare the trajectory predictors, we use 
the \emph{Top-$K$ Average} and \emph{Final Displacement Errors} (Top-$K$ ADE and FDE, in pixels for SDD and meters for TH\"{O}R-MAGNI), 
as in~\cite{salzmann20,kothari23}. Top-$K$ ADE measures the average $\ell_2$ distance between the ground truth track and the closest prediction (out of $K$ samples), and FDE measures the distance between the last predicted position and the corresponding  ground truth. 
We present the results of Top-$1$ and Top-$3$ ADE/FDE. When $K=1$, we use the most likely output trajectory. 
We measure the mean and standard deviation of these metrics across iterations in the validation.

\section{RESULTS} \label{section-results}

\begin{table*}[!t]
    \centering
    \caption{Top-1 ADE/FDE scores (ADE above, FDE below) in TH\"{O}R-MAGNI Scenario 2 and SDD datasets with a 90\% train ratio. Bold values highlight superior performance of conditional models over their unconditional counterparts across most settings.}
    \label{tab:t1_ade_fde}
    \resizebox{\linewidth}{!}{
        \begin{tabular}{@{}c|c|cc|cc|cc|cc|cc@{}}
        \toprule
        \textbf{\begin{tabular}[c]{@{}c@{}}Data\end{tabular}} &
            \textbf{Class} &
            \textbf{RED} &
            \textbf{cRED} &
            \textbf{TF} &
            \textbf{cTF} &
            \textbf{GAN} &
            \textbf{cGAN} &
            \textbf{VAE} &
            \textbf{cVAE} &
            \textbf{MoD} &
            \textbf{cMoD} \\ \midrule
        \multirow{13}{*}{\rotatebox{90}{TH\"{O}R-MAGNI Scenario~2}} &
            \begin{tabular}[c]{@{}c@{}}\emph{Carrier-}\\ \emph{Box}\end{tabular} &
            \begin{tabular}[c]{@{}c@{}}0.64$\pm$0.07\\ 1.23$\pm$0.14\end{tabular} &
            \textbf{\begin{tabular}[c]{@{}c@{}}0.60$\pm$0.07\\ 1.10$\pm$0.14\end{tabular}} &
            \begin{tabular}[c]{@{}c@{}}0.66$\pm$0.07\\ 1.24$\pm$0.15\end{tabular} &
            \textbf{\begin{tabular}[c]{@{}c@{}}0.60$\pm$0.07\\ 1.10$\pm$0.13\end{tabular}} &
            \begin{tabular}[c]{@{}c@{}}0.76$\pm$0.07\\ 1.50$\pm$0.19\end{tabular} &
            \textbf{\begin{tabular}[c]{@{}c@{}}0.70$\pm$0.10\\ 1.33$\pm$0.19\end{tabular}} &
            \begin{tabular}[c]{@{}c@{}}0.68$\pm$0.06\\ 1.31$\pm$0.13\end{tabular} &
            \textbf{\begin{tabular}[c]{@{}c@{}}0.66$\pm$0.05\\ 1.26$\pm$0.11\end{tabular}} &
            \begin{tabular}[c]{@{}c@{}}0.81$\pm$0.11\\ 1.59$\pm$0.25\end{tabular} &
            \textbf{\begin{tabular}[c]{@{}c@{}}0.73$\pm$0.07\\ 1.40$\pm$0.17\end{tabular}}
            \\ \cmidrule(l){2-12} 
            &
            \begin{tabular}[c]{@{}c@{}}\emph{Carrier-}\\ \emph{Bucket}\end{tabular} &
            \begin{tabular}[c]{@{}c@{}}0.71$\pm$0.06\\ 1.35$\pm$0.18\end{tabular} &
            \textbf{\begin{tabular}[c]{@{}c@{}}0.67$\pm$0.06\\ 1.21$\pm$0.15\end{tabular}} &
            \begin{tabular}[c]{@{}c@{}}0.65$\pm$0.05\\ 1.24$\pm$0.13\end{tabular} &
            \textbf{\begin{tabular}[c]{@{}c@{}}0.60$\pm$0.06\\ 1.12$\pm$0.16\end{tabular}} &
            \begin{tabular}[c]{@{}c@{}}0.78$\pm$0.04\\ 1.48$\pm$0.18\end{tabular} &
            \textbf{\begin{tabular}[c]{@{}c@{}}0.73$\pm$0.06\\ 1.37$\pm$0.14\end{tabular}} &
            \begin{tabular}[c]{@{}c@{}}\textbf{0.73}$\pm$\textbf{0.09}\\ 1.44$\pm$0.20\end{tabular} &
            \begin{tabular}[c]{@{}c@{}}0.74$\pm$0.08\\ \textbf{1.43}$\pm$\textbf{0.19}\end{tabular} &
            \begin{tabular}[c]{@{}c@{}}0.92$\pm$0.18\\ 1.78$\pm$0.37\end{tabular} &
            \textbf{\begin{tabular}[c]{@{}c@{}}0.72$\pm$0.10\\ 1.30$\pm$0.17\end{tabular}}
            \\ \cmidrule(l){2-12} 
            &
            \begin{tabular}[c]{@{}c@{}}\emph{Visitors-}\\ \emph{Alone}\end{tabular} &
            \begin{tabular}[c]{@{}c@{}}0.81$\pm$0.05\\ 1.53$\pm$0.12\end{tabular} &
            \textbf{\begin{tabular}[c]{@{}c@{}}0.78$\pm$0.06\\ 1.48$\pm$0.13\end{tabular}} &
            \begin{tabular}[c]{@{}c@{}}0.79$\pm$0.04\\ 1.52$\pm$0.12\end{tabular} &
            \textbf{\begin{tabular}[c]{@{}c@{}}0.75$\pm$0.04\\ 1.45$\pm$0.14\end{tabular}} &
            \begin{tabular}[c]{@{}c@{}}0.88$\pm$0.07\\ 1.72$\pm$0.14\end{tabular} &
            \textbf{\begin{tabular}[c]{@{}c@{}}0.85$\pm$0.08\\ 1.67$\pm$0.19\end{tabular}} &
            \begin{tabular}[c]{@{}c@{}}0.84$\pm$0.06\\ 1.62$\pm$0.16\end{tabular} &
            \textbf{\begin{tabular}[c]{@{}c@{}}0.83$\pm$0.05\\ \textbf{1.61}$\pm$\textbf{0.14}\end{tabular}} &
            \begin{tabular}[c]{@{}c@{}}0.94$\pm$0.06\\ 1.97$\pm$0.20\end{tabular} &
            \textbf{\begin{tabular}[c]{@{}c@{}}0.92$\pm$0.09\\ 1.95$\pm$0.22\end{tabular}}
            \\ \cmidrule(l){2-12} 
            &
            \begin{tabular}[c]{@{}c@{}}\emph{Visitors-}\\ \emph{Group}\end{tabular} &
            \textbf{\begin{tabular}[c]{@{}c@{}}0.72$\pm$0.05\\ 1.34$\pm$0.17\end{tabular}} &
            \begin{tabular}[c]{@{}c@{}}\textbf{0.72}$\pm$\textbf{0.07}\\ 1.35$\pm$0.18\end{tabular} &
            \begin{tabular}[c]{@{}c@{}}0.74$\pm$0.06\\ 1.40$\pm$0.15\end{tabular} &
            \textbf{\begin{tabular}[c]{@{}c@{}}0.68$\pm$0.05\\ 1.29$\pm$0.13\end{tabular}} &
            \textbf{\begin{tabular}[c]{@{}c@{}}0.80$\pm$0.08\\ 1.52$\pm$0.15\end{tabular}} &
            \begin{tabular}[c]{@{}c@{}}\textbf{0.80}$\pm$\textbf{0.06}\\ 1.57$\pm$0.16\end{tabular} &
            \begin{tabular}[c]{@{}c@{}}0.78$\pm$0.08\\ 1.59$\pm$0.16\end{tabular} &
            \textbf{\begin{tabular}[c]{@{}c@{}}0.75$\pm$0.06\\ 1.56$\pm$0.13\end{tabular}} &
            \textbf{\begin{tabular}[c]{@{}c@{}}0.82$\pm$0.10\\ 1.78$\pm$0.24\end{tabular}} &
            \begin{tabular}[c]{@{}c@{}}0.83$\pm$0.10\\ 1.80$\pm$0.24\end{tabular}
            \\ \cmidrule(l){2-12} 
            &
            \begin{tabular}[c]{@{}c@{}}\emph{Carrier-}\\ \emph{LO}\end{tabular} &
            \begin{tabular}[c]{@{}c@{}}0.73$\pm$0.05\\ 1.44$\pm$0.12\end{tabular} &
            \textbf{\begin{tabular}[c]{@{}c@{}}0.69$\pm$0.03\\ 1.38$\pm$0.10\end{tabular}} &
            \begin{tabular}[c]{@{}c@{}}0.69$\pm$0.05\\ 1.41$\pm$0.12\end{tabular} &
            \textbf{\begin{tabular}[c]{@{}c@{}}0.64$\pm$0.04\\ 1.31$\pm$0.08\end{tabular}} &
            \begin{tabular}[c]{@{}c@{}}0.78$\pm$0.08\\ 1.59$\pm$0.16\end{tabular} &
            \textbf{\begin{tabular}[c]{@{}c@{}}0.75$\pm$0.06\\ 1.56$\pm$0.13\end{tabular}} &
            \begin{tabular}[c]{@{}c@{}}0.77$\pm$0.07\\ 1.50$\pm$0.15\end{tabular} &
            \textbf{\begin{tabular}[c]{@{}c@{}}0.72$\pm$0.06\\ 1.44$\pm$0.12\end{tabular}} &
            \begin{tabular}[c]{@{}c@{}}0.83$\pm$0.10\\ 1.73$\pm$0.22\end{tabular} &
            \textbf{\begin{tabular}[c]{@{}c@{}}0.75$\pm$0.08\\ 1.61$\pm$0.19\end{tabular}}
            \\ \cmidrule(l){2-12}
            &
            \textbf{Global} &
            \begin{tabular}[c]{@{}c@{}}0.74$\pm$0.01\\ 1.41$\pm$0.04\end{tabular} &
            \textbf{\begin{tabular}[c]{@{}c@{}}0.71$\pm$0.03\\ 1.34$\pm$0.06\end{tabular}} &
            \begin{tabular}[c]{@{}c@{}}0.72$\pm$0.02\\ 1.39$\pm$0.04\end{tabular} &
            \textbf{\begin{tabular}[c]{@{}c@{}}0.67$\pm$0.02\\ 1.29$\pm$0.05\end{tabular}} &
            \begin{tabular}[c]{@{}c@{}}0.81$\pm$0.05\\ 1.59$\pm$0.09\end{tabular} &
            \textbf{\begin{tabular}[c]{@{}c@{}}0.78$\pm$0.05\\ 1.53$\pm$0.09\end{tabular}} &
            \begin{tabular}[c]{@{}c@{}}0.77$\pm$0.03\\ 1.49$\pm$0.05\end{tabular} &
            \textbf{\begin{tabular}[c]{@{}c@{}}0.76$\pm$0.02\\ 1.47$\pm$0.06\end{tabular}} &
            \begin{tabular}[c]{@{}c@{}}0.87$\pm$0.05 \\ 1.79$\pm$0.10\end{tabular} &
            \textbf{\begin{tabular}[c]{@{}c@{}}0.80$\pm$0.05 \\ 1.67$\pm$0.09\end{tabular}} \\ \specialrule{.2em}{.1em}{.1em}
        \multirow{9}{*}{\rotatebox{90}{SDD}} &
            \emph{Ped.} &
            \textbf{\begin{tabular}[c]{@{}c@{}}18.63$\pm$0.54\\ 37.55$\pm$1.22\end{tabular}} &
            \begin{tabular}[c]{@{}c@{}}18.76$\pm$0.54\\ 37.69$\pm$1.08\end{tabular} &
            \textbf{\begin{tabular}[c]{@{}c@{}}18.99$\pm$0.89\\ 37.60$\pm$1.76\end{tabular}} &
            \begin{tabular}[c]{@{}c@{}}19.00$\pm$0.79\\ 37.72$\pm$1.30\end{tabular} &
            \begin{tabular}[c]{@{}c@{}}20.26$\pm$0.69\\ \textbf{40.27}$\pm$\textbf{1.25}\end{tabular} &
            \begin{tabular}[c]{@{}c@{}}\textbf{20.21}$\pm$\textbf{0.49}\\ 40.31$\pm$0.92\end{tabular} &
            \begin{tabular}[c]{@{}c@{}}20.92$\pm$1.25\\ 41.49$\pm$2.32\end{tabular} &
            \textbf{\begin{tabular}[c]{@{}c@{}}20.09$\pm$0.77\\ 39.90$\pm$1.44\end{tabular}} &
            \begin{tabular}[c]{@{}c@{}}19.88$\pm$0.46\\ 40.02$\pm$1.07\end{tabular} &
            \textbf{\begin{tabular}[c]{@{}c@{}}19.69$\pm$0.46\\ 39.64$\pm$1.14\end{tabular}} \\ \cmidrule(l){2-12} 
            &
            \emph{Car} &
            \begin{tabular}[c]{@{}c@{}}\textbf{8.44}$\pm$\textbf{7.48}\\ 16.63$\pm$15.03\end{tabular} &
            \begin{tabular}[c]{@{}c@{}}8.66$\pm$7.05\\ \textbf{16.55}$\pm$\textbf{14.30}\end{tabular} &
            \textbf{\begin{tabular}[c]{@{}c@{}}9.36$\pm$6.91\\ 17.79$\pm$14.36\end{tabular}} &
            \begin{tabular}[c]{@{}c@{}}9.64$\pm$7.62\\ 18.11$\pm$15.32\end{tabular} &
            \begin{tabular}[c]{@{}c@{}}10.99$\pm$7.39\\ 20.78$\pm$14.83\end{tabular} &
            \textbf{\begin{tabular}[c]{@{}c@{}}10.51$\pm$7.12\\ 19.55$\pm$14.23\end{tabular}} &
            \begin{tabular}[c]{@{}c@{}}10.83$\pm$7.00\\ 20.81$\pm$13.94\end{tabular} &
            \textbf{\begin{tabular}[c]{@{}c@{}}10.41$\pm$7.28\\ 18.72$\pm$13.86\end{tabular}} &
            \begin{tabular}[c]{@{}c@{}}9.95$\pm$11.05 \\ 20.61$\pm$23.07\end{tabular} &
            \textbf{\begin{tabular}[c]{@{}c@{}}8.73$\pm$9.56\\ 18.48$\pm$20.16\end{tabular}} \\ \cmidrule(l){2-12} 
            &
            \emph{Byc.} &
            \textbf{\begin{tabular}[c]{@{}c@{}}64.08$\pm$2.49\\ 137.42$\pm$5.31\end{tabular}} &
            \begin{tabular}[c]{@{}c@{}}64.38$\pm$2.53\\ 137.56$\pm$4.97\end{tabular} &
            \begin{tabular}[c]{@{}c@{}}65.33$\pm$2.37\\ 142.08$\pm$4.60\end{tabular} &
            \textbf{\begin{tabular}[c]{@{}c@{}}64.01$\pm$2.67\\ 139.97$\pm$4.75\end{tabular}} &
            \begin{tabular}[c]{@{}c@{}}67.32$\pm$2.75\\ 145.50$\pm$5.04\end{tabular} &
            \textbf{\begin{tabular}[c]{@{}c@{}}67.04$\pm$2.41\\ 144.50$\pm$4.61\end{tabular}} &
            \begin{tabular}[c]{@{}c@{}}68.22$\pm$3.58\\ 147.00$\pm$6.32\end{tabular} &
            \textbf{\begin{tabular}[c]{@{}c@{}}67.80$\pm$3.36\\ 145.13$\pm$6.64\end{tabular}} &
            \begin{tabular}[c]{@{}c@{}}64.35$\pm$2.02\\ 142.51$\pm$4.40\end{tabular} &
            \textbf{\begin{tabular}[c]{@{}c@{}}63.60$\pm$2.02\\ 141.01$\pm$4.24\end{tabular}} \\ \cmidrule(l){2-12} 
            &
            \textbf{Global} &
            \textbf{\begin{tabular}[c]{@{}c@{}}33.95$\pm$0.91\\ 71.23$\pm$2.07\end{tabular}} &
            \begin{tabular}[c]{@{}c@{}}34.14$\pm$1.00\\ 71.38$\pm$1.99\end{tabular} &
            \begin{tabular}[c]{@{}c@{}}34.62$\pm$0.90\\ 72.87$\pm$1.70\end{tabular} &
            \textbf{\begin{tabular}[c]{@{}c@{}}34.19$\pm$1.00\\ 72.23$\pm$1.63\end{tabular}} &
            \begin{tabular}[c]{@{}c@{}}36.14$\pm$1.07\\ 75.79$\pm$1.96\end{tabular} &
            \textbf{\begin{tabular}[c]{@{}c@{}}36.01$\pm$0.75\\ 75.46$\pm$1.55\end{tabular}} &
            \begin{tabular}[c]{@{}c@{}}36.87$\pm$1.67\\ 77.09$\pm$3.08\end{tabular} &
            \textbf{\begin{tabular}[c]{@{}c@{}}36.19$\pm$1.45\\ 75.40$\pm$2.81\end{tabular}} &
            \begin{tabular}[c]{@{}c@{}}34.60$\pm$0.62 \\ 74.03$\pm$1.50\end{tabular} &
            \textbf{\begin{tabular}[c]{@{}c@{}}34.21$\pm$0.73 \\ 73.25$\pm$1.60\end{tabular}} \\ \bottomrule
        \end{tabular}%
    }
    \vspace{-2mm}
\end{table*}

\pgfplotstableread[col  sep=comma]{contents/results_tab/magni/Scenario_2_top1_global_ade_latex.csv}\sctwoadetopone
\pgfplotstableread[col sep=comma]{contents/results_tab/magni/Scenario_2_top1_global_fde_latex.csv}\sctwofdetopone
\pgfplotstableread[col  sep=comma]{contents/results_tab/magni/Scenario_2_top3_global_ade_latex.csv}\sctwoadetopthree
\pgfplotstableread[col sep=comma]{contents/results_tab/magni/Scenario_2_top3_global_fde_latex.csv}\sctwofdetopthree

\pgfplotstableread[col  sep=comma]{contents/results_tab/magni/Scenario_3A_top1_global_ade_latex.csv}\sctaadetopone
\pgfplotstableread[col sep=comma]{contents/results_tab/magni/Scenario_3A_top1_global_fde_latex.csv}\sctafdetopone
\pgfplotstableread[col  sep=comma]{contents/results_tab/magni/Scenario_3A_top3_global_ade_latex.csv}\sctaadetopthree
\pgfplotstableread[col sep=comma]{contents/results_tab/magni/Scenario_3A_top3_global_fde_latex.csv}\sctafdetopthree

\pgfplotstableread[col  sep=comma]{contents/results_tab/magni/Scenario_3B_top1_global_ade_latex.csv}\sctbadetopone
\pgfplotstableread[col sep=comma]{contents/results_tab/magni/Scenario_3B_top1_global_fde_latex.csv}\sctbfdetopone
\pgfplotstableread[col  sep=comma]{contents/results_tab/magni/Scenario_3B_top3_global_ade_latex.csv}\sctbadetopthree
\pgfplotstableread[col sep=comma]{contents/results_tab/magni/Scenario_3B_top3_global_fde_latex.csv}\sctbfdetopthree

\pgfplotstableread[col  sep=comma]{contents/results_tab/sdd/sdd_top1_global_ade_latex.csv}\sddadetopone
\pgfplotstableread[col sep=comma]{contents/results_tab/sdd/sdd_top1_global_fde_latex.csv}\sddfdetopone
\pgfplotstableread[col  sep=comma]{contents/results_tab/sdd/sdd_top3_global_ade_latex.csv}\sddadetopthree
\pgfplotstableread[col sep=comma]{contents/results_tab/sdd/sdd_top3_global_fde_latex.csv}\sddfdetopthree

\begin{figure*}[t]
\centering
\includegraphics[width=12mm, angle=90]{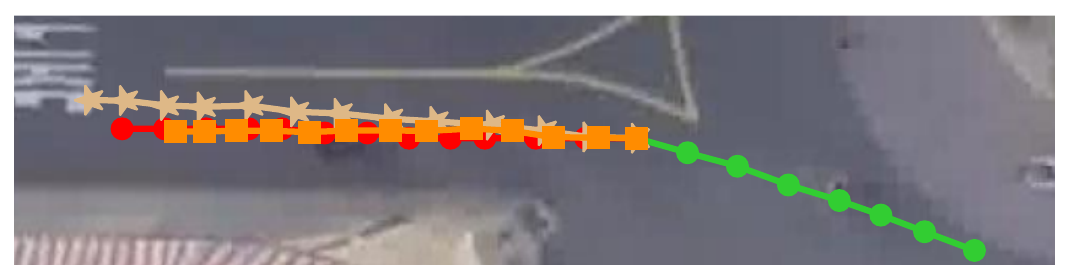}
\includegraphics[width=12mm, angle=90]{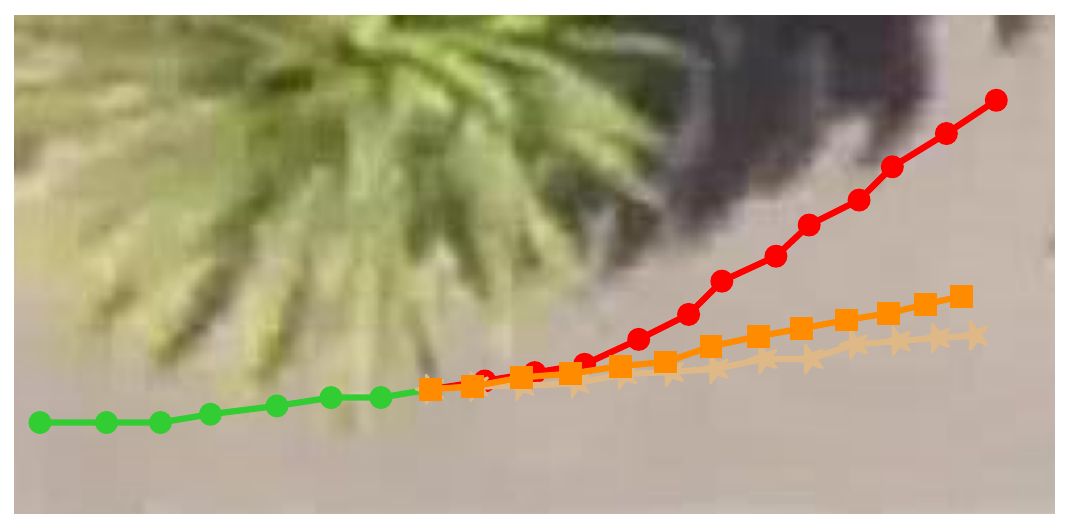}
\includegraphics[height=12mm]{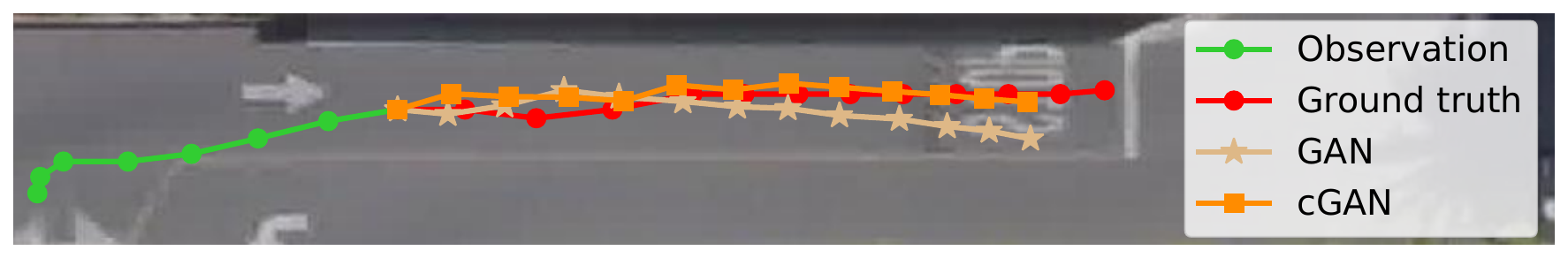}

\includegraphics[width=12mm, angle=90]{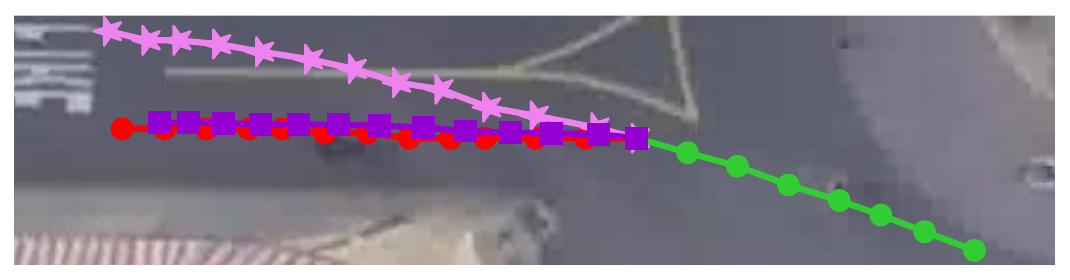}
\includegraphics[width=12mm, angle=90]{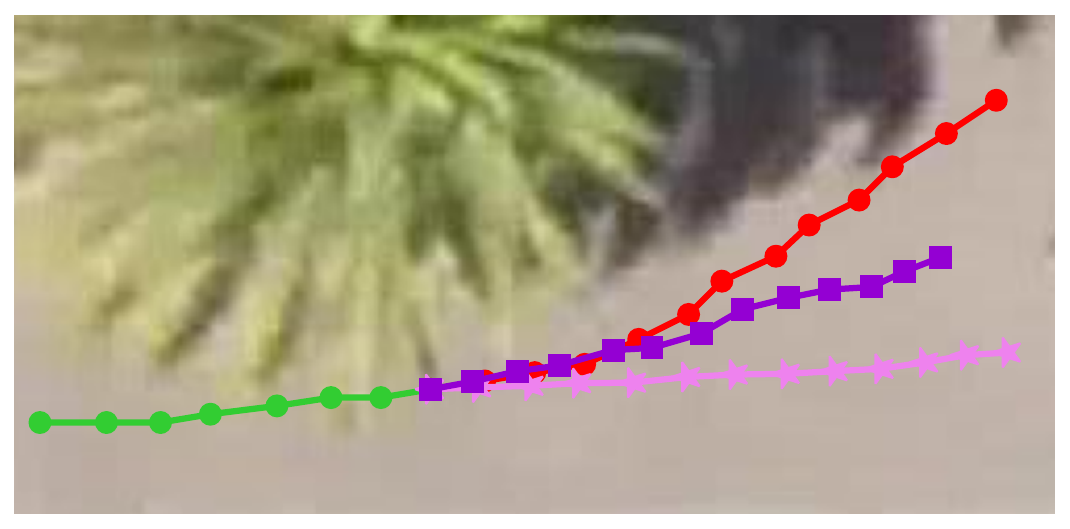}
\includegraphics[height=12mm]{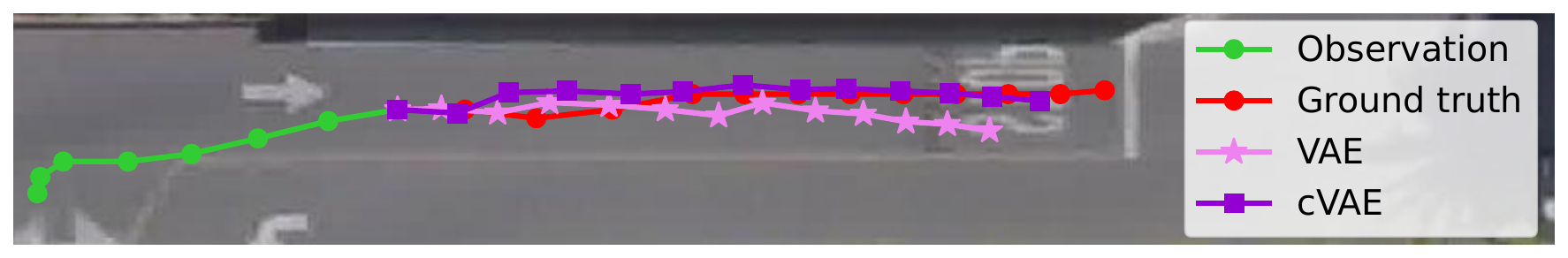}

\includegraphics[width=12mm, angle=90]{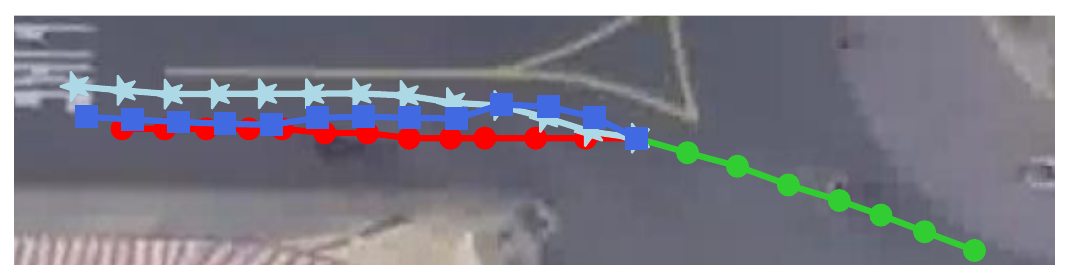}
\includegraphics[width=12mm, angle=90]{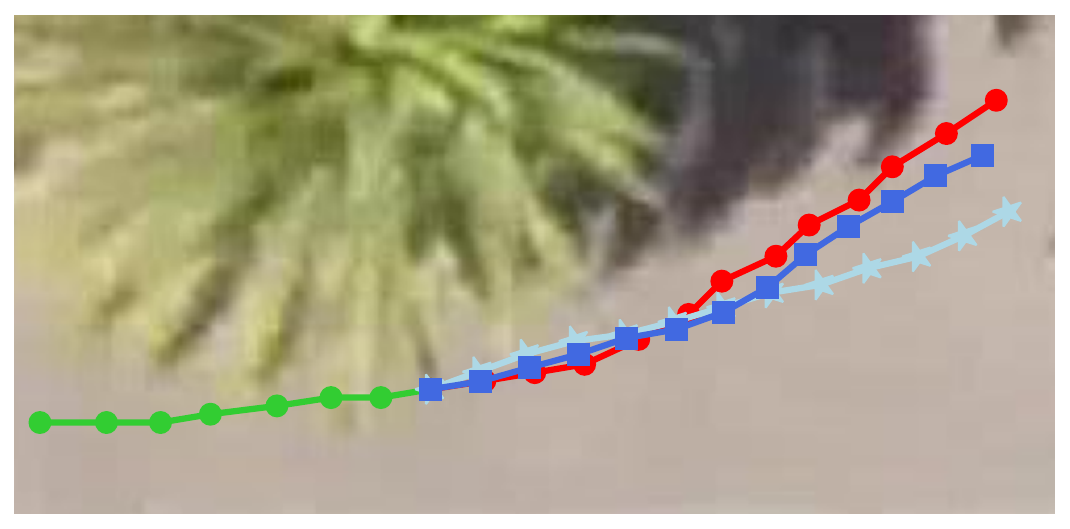}
\includegraphics[height=12mm]{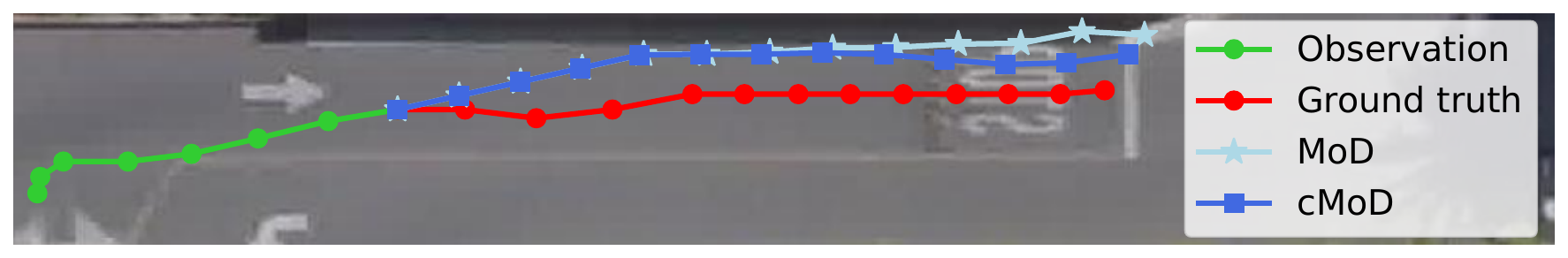}
\caption{Prediction examples of {\em Byciclist} (\textbf{left}), {\em Pedestrian} (\textbf{middle}) and {\em Car} (\textbf{right}) in SDD with \SI{4.8}{\second} prediction horizon.}
\label{fig:sdd_quality_res}
\vspace*{-4mm}
\end{figure*}

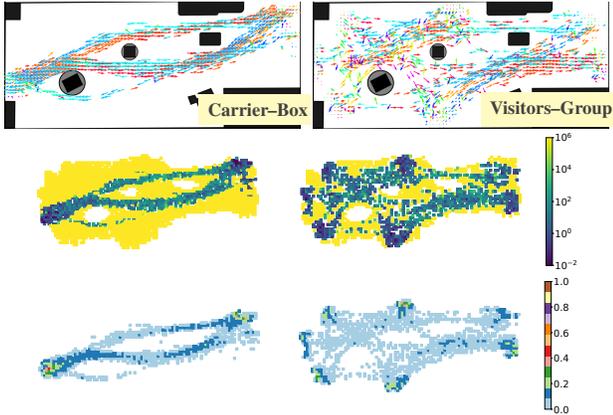
\begin{figure}
\centering

\begin{tikzpicture}
  \node[anchor=south west,inner sep=0] (image1) at (0,0) {\includegraphics[width=.45\linewidth]{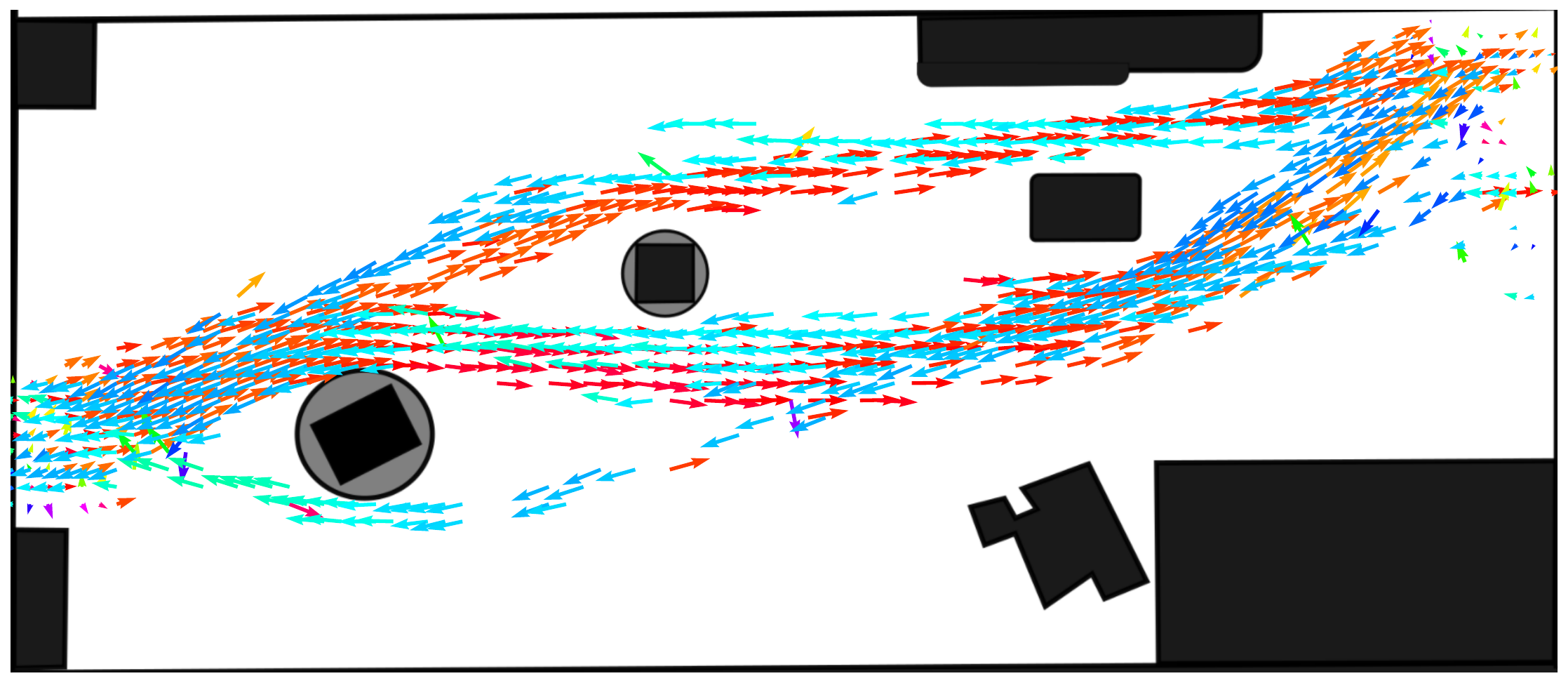}};
  \node[font=\bfseries\scriptsize, text=darkgray, fill=yellow!30] at  (image1.center) [xshift=14mm, yshift=-6mm] {Carrier--Box}; 

  \node[anchor=south west,inner sep=0, right=1mm of image1] (image2) {\includegraphics[width=.45\linewidth]{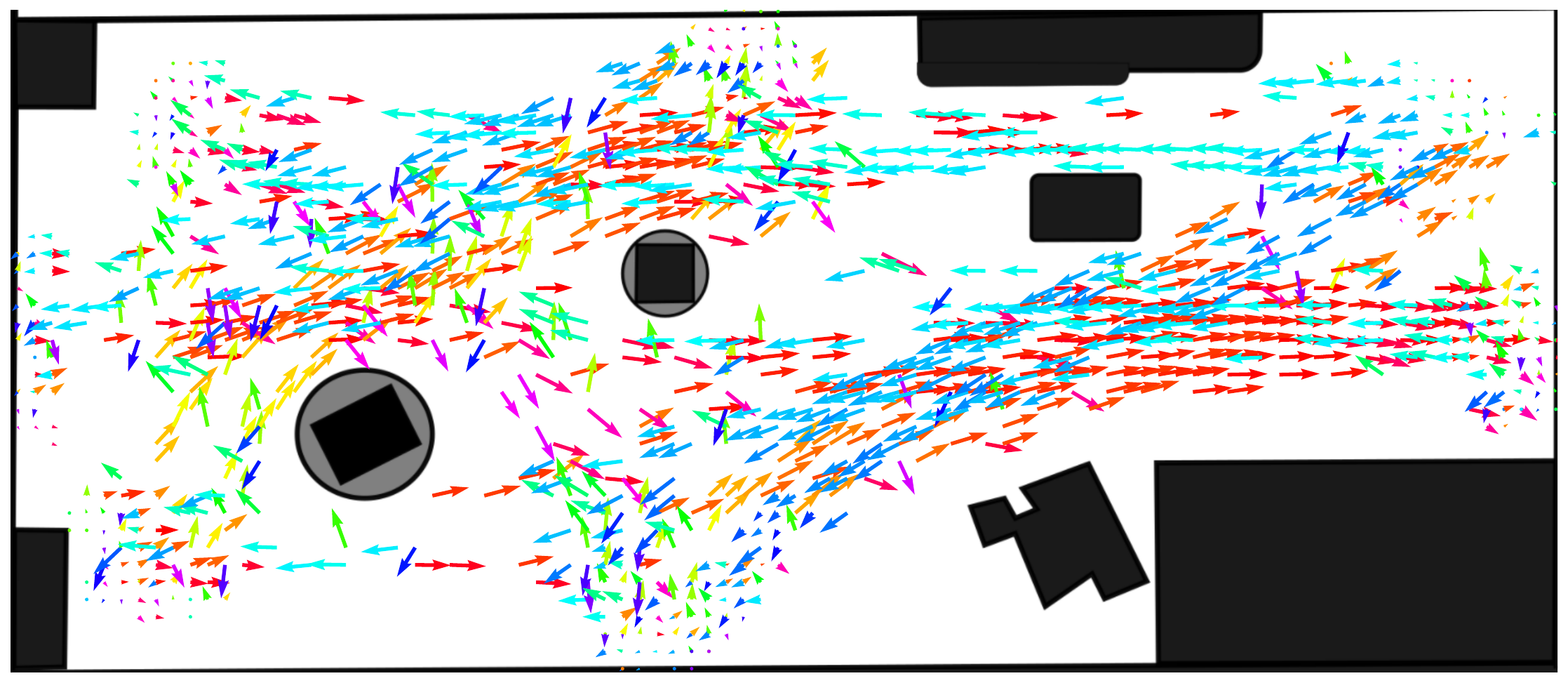}};
  \node[font=\bfseries\scriptsize, text=darkgray, fill=yellow!30] at  (image2.center) [xshift=12mm, yshift=-6mm] {Visitors--Group}; 
  
  \node[anchor=south west,inner sep=0, below=0mm of image1] (image3) {\includegraphics[clip,trim=0mm 0mm 35mm 0mm, height=19mm]{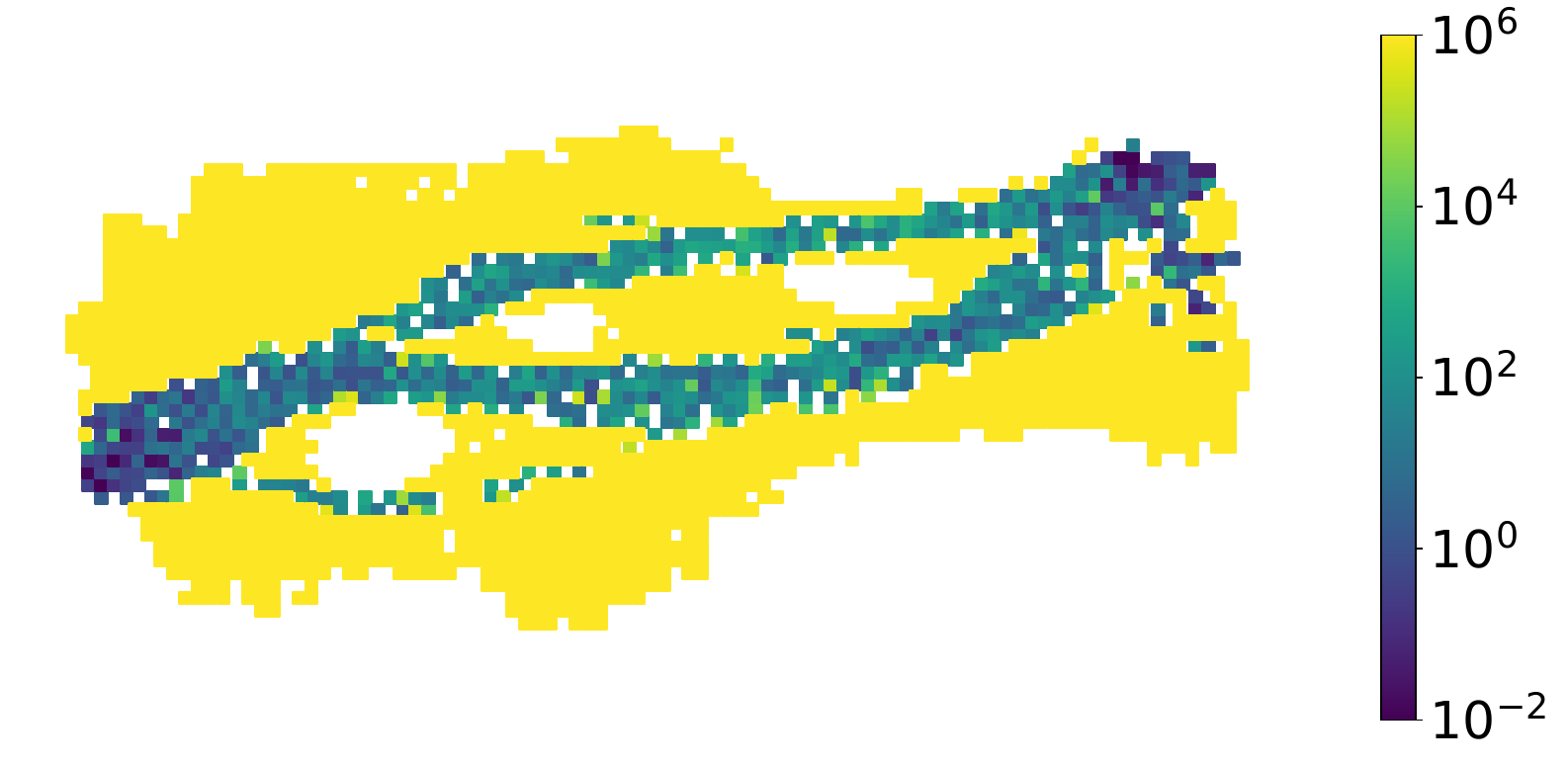}};
  \node[font=\bfseries\scriptsize, text=black] at (image3.center) [] {}; 

  \node[anchor=south west,inner sep=0, right=1mm of image3] (image4) {\includegraphics[clip,trim=0mm 0mm 0mm 0mm, height=19mm]{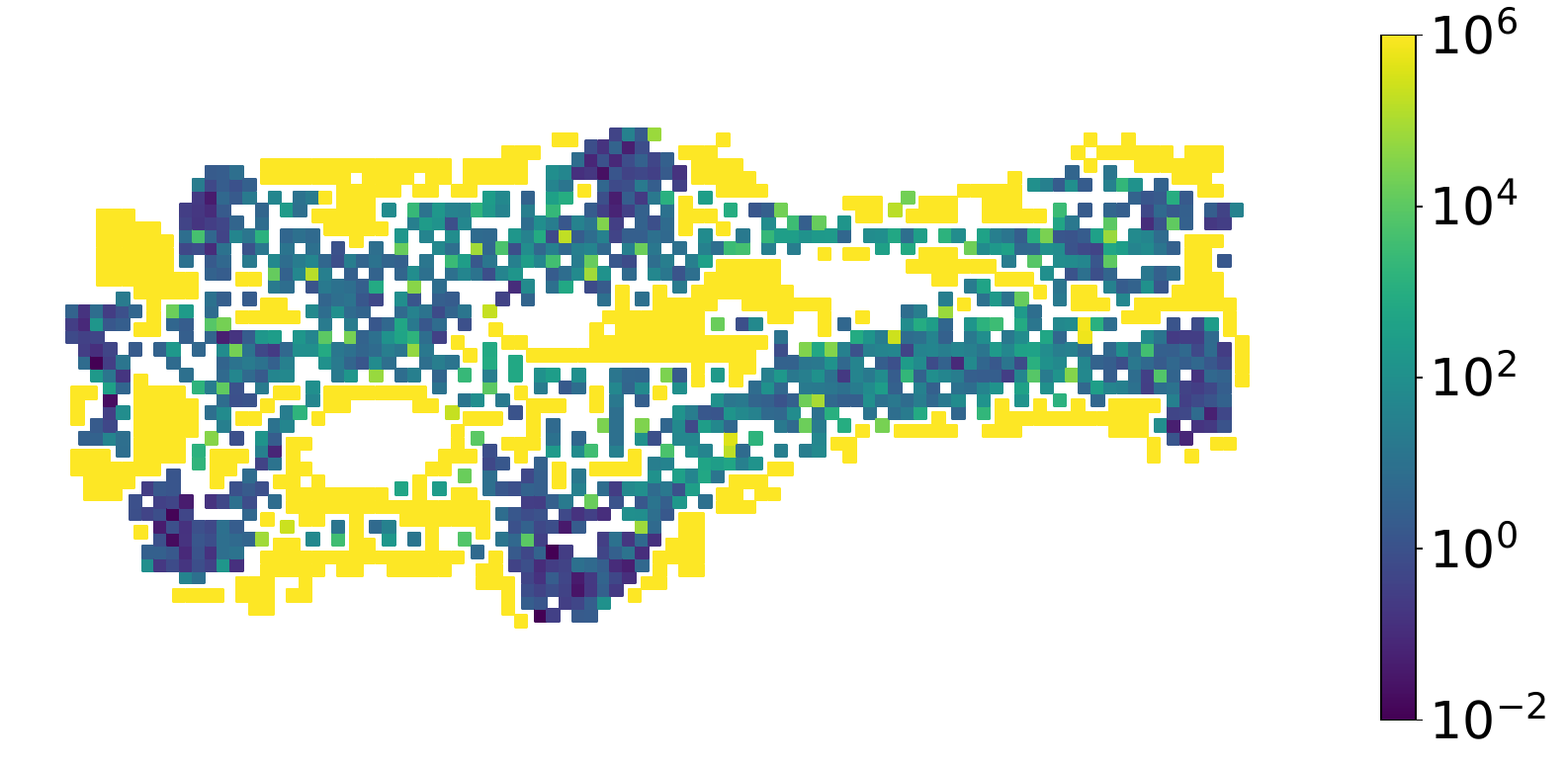}};
  \node[font=\bfseries\scriptsize, text=black] at (image4.center) [] {}; 

  \node[anchor=south west,inner sep=0, below=0mm of image3] (image5) {\includegraphics[clip,trim=0mm 0mm 30mm 0mm, height=19mm]{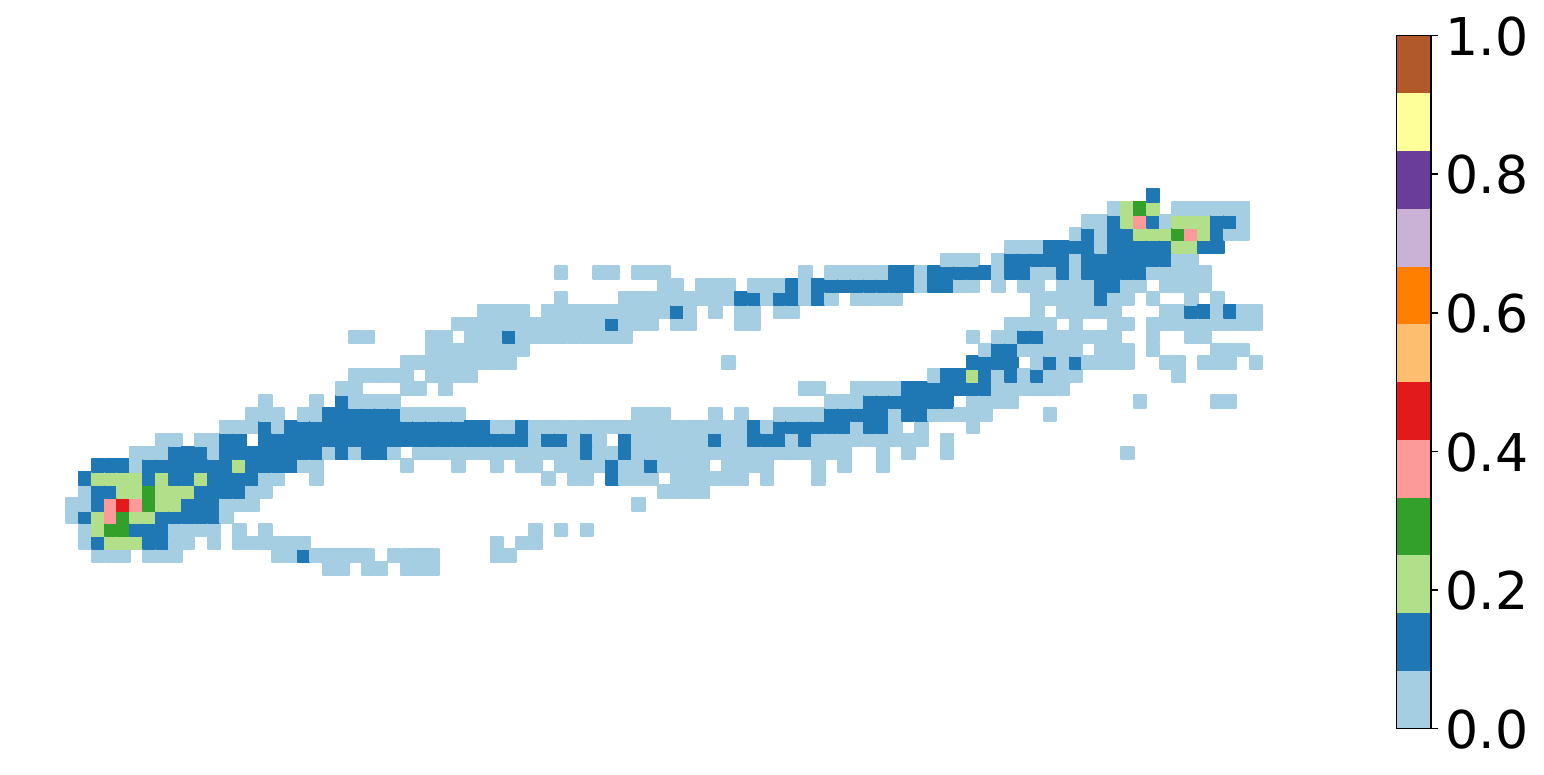}};
  \node[font=\bfseries\scriptsize, text=black] at (image5.center) [] {}; 
  
  \node[anchor=south west,inner sep=0, right=1mm of image5] (image6) {\includegraphics[clip,trim=0mm 0mm 0mm 0mm, height=19mm]{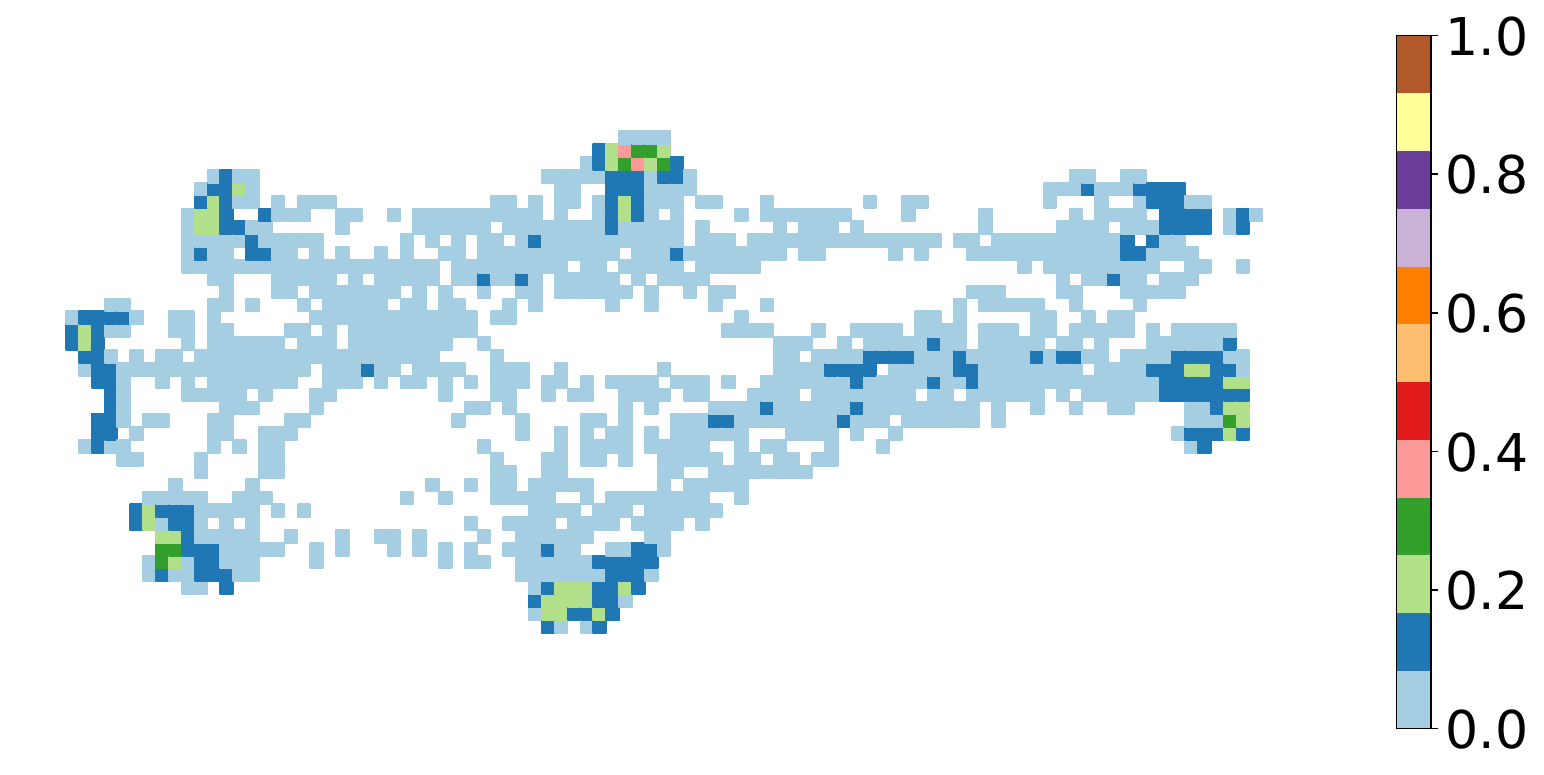}};
  \node[font=\bfseries\scriptsize, text=black] at  (image6.center) [] {}; 
\end{tikzpicture}

\vspace*{-2mm}
\caption{
Comparison of motion patterns of {\em Carrier--Box} and {\em Visitors--Group} in TH\"{O}R-MAGNI, Scenario~2. Class-conditioned CLiFF-maps (\textbf{first row}) show that {\em Carrier--Box} has a more distinct and structured motion pattern compared to {\em Visitors--Group}. The KL divergence heatmap (\textbf{middle row}) quantifies the difference between the class-conditioned CLiFF-map and the general one. {\em Visitors--Group} shows less divergence from the general motion patterns and lower motion intensity (\textbf{bottom row}), resulting in a less pronounced improvement in prediction accuracy from using class labels.
}
\label{fig:CLiFF-mapMAGNI}
\vspace*{-6mm}
\end{figure}

In our analysis, we aim to (1) quantify the improvement in trajectory prediction performance when using class attributes, and (2) evaluate trajectory prediction performance based on the specific characteristics of the dataset. The latter provides insight into the appropriate trajectory prediction method selection for a particular application.
Due to space limitations, we only report the results for TH\"{O}R-MAGNI Scenario~2 in \cref{tab:t1_ade_fde}, \cref{fig:CLiFF-mapMAGNI} and \cref{fig:magni_quality_res}, and for Scenarios 3A and 3B in \cref{fig:det_deep_learning_sc3ab}. We observe similar trends in all scenarios.

\subsection{Accuracy Analysis Conditioned on Class Balance}\label{sec:acc}
\cref{tab:t1_ade_fde} shows the prediction accuracy results separately for each class on Scenario~2 of the TH\"{O}R-MAGNI and SDD datasets. It also shows the global results for all trajectories (last rows for each dataset).
A broad view of the TH\"{O}R-MAGNI results shows that conditional methods outperform their unconditional counterparts regardless of the
type of method (deep learning and MoD). When predicting trajectories from the {\em Visitors--Group}, this difference 
is least pronounced. We speculate that this may be due to the fact that the motion patterns of these agents 
are less structured compared to the other classes, as shown in~\cref{fig:CLiFF-mapMAGNI}. This also highlights the importance of suitable class labels, such that each class encompasses specific motion patterns, and dataset-imposed classes may not always do so.
For the imbalanced dataset (SDD), deep learning 
methods face the challenge of identifying a representative number of different motion patterns across classes. This difficulty is most pronounced in single-output deep learning methods (RED and TF). In contrast, cMoD is less sensitive to class proportions and is able to use 
class information for more accurate predictions. 
In summary, we highlight two key points: (1) the superiority of deep learning methods over MoD-based approaches in balanced 
datasets like TH\"{O}R-MAGNI, and (2) the appropriateness of conditional MoD over deep generative methods (cGAN, cVAE) 
for imbalanced datasets like SDD.

In the MoD-aware predictor, cMoD outperforms general MoD in both datasets. The TH\"{O}R-MAGNI dataset highlights differences in spatial patterns among classes, as shown in \cref{fig:CLiFF-mapMAGNI}. 
Prediction accuracy improvements were more pronounced in classes with distinct motion patterns, such as {\em Carrier--Box} and {\em Carrier--Bucket}, which deviate more from the general motion pattern.
In SDD, variances in speed are observed among different classes, as depicted in \cref{fig:CLiFF-mapSDD}. 
A single CLiFF-map struggles to accurately model variations across multiple classes, leading to inaccurate predictions compared to the class-conditioned MoD-aware method.

\subsection{Data Efficiency Analysis}\label{sec:data_eff}
\begin{figure}[!t]
    \centering
    \resizebox{.49\columnwidth}{!}{
    \begin{tikzpicture}
        \begin{axis}[
            ylabel={Top-$1$ ADE (m)},
            label style={font=\Large},
            tick label style={font=\Large},
            legend columns=2,
            grid=major,
            ymax=1.2,
            xtick=data,
            xticklabels={},
            legend style={nodes={scale=1.7, transform shape}}, 
        ]
        
        \addplot[magenta, mark=diamond, mark size=4pt]
        plot [error bars/.cd, y dir=both, y explicit]
        table[x expr={100 - \thisrow{test_ratio}}, y=LSTM_avg]{\sctaadetopone};
        \addlegendentryexpanded{RED};

        \addplot[magenta, mark=asterisk, mark size=4pt] 
        plot [error bars/.cd, y dir=both, y explicit]
        table[x expr={100 - \thisrow{test_ratio}}, y=cLSTM_avg]{\sctaadetopone};
        \addlegendentryexpanded{cRED};

        \addplot[black, mark=oplus, mark size=4pt]
        plot [error bars/.cd, y dir=both, y explicit]
        table[x expr={100 - \thisrow{test_ratio}}, y=TF_avg]{\sctaadetopone};
        \addlegendentryexpanded{TF};

        \addplot[black, mark=triangle, mark size=4pt]
        plot [error bars/.cd, y dir=both, y explicit]
        table[x expr={100 - \thisrow{test_ratio}}, y=cTF_avg]{\sctaadetopone};
        \addlegendentryexpanded{cTF};

        \addplot[teal,mark=square, mark size=4pt]
        plot [error bars/.cd, y dir=both, y explicit]
        table[x expr={100 - \thisrow{test_ratio}}, y=MoD_avg]{\sctaadetopone};
        \addlegendentryexpanded{MoD};

        \addplot[teal,mark=|, mark size=4pt]
        plot [error bars/.cd, y dir=both, y explicit]
        table[x expr={100 - \thisrow{test_ratio}}, y=cMoD_avg]{\sctaadetopone};
        \addlegendentryexpanded{cMoD};
        
        \end{axis}
    \end{tikzpicture}
    }
    \resizebox{.49\columnwidth}{!}{
        \begin{tikzpicture}
            \begin{axis}[
                ylabel={Top-$1$ FDE (m)},
                label style={font=\Large},
                tick label style={font=\Large},
                legend columns=2,
                grid=major,
                ymax=2.50,
                xtick=data,
                xticklabels={},
                legend style={nodes={scale=1.7, transform shape}}, 
            ]
            
            \addplot[magenta, mark=diamond, mark size=4pt]
            plot [error bars/.cd, y dir=both, y explicit]
            table[x expr={100 - \thisrow{test_ratio}}, y=LSTM_avg]{\sctafdetopone};
            \addlegendentryexpanded{RED};

            \addplot[magenta, mark=asterisk, mark size=4pt]
            plot [error bars/.cd, y dir=both, y explicit]
            table[x expr={100 - \thisrow{test_ratio}}, y=cLSTM_avg]{\sctafdetopone};
            \addlegendentryexpanded{cRED};
    
            \addplot[black, mark=oplus, mark size=4pt]
            plot [error bars/.cd, y dir=both, y explicit]
            table[x expr={100 - \thisrow{test_ratio}}, y=TF_avg]{\sctafdetopone};
            \addlegendentryexpanded{TF};

            \addplot[black, mark=triangle, mark size=4pt]
            plot [error bars/.cd, y dir=both, y explicit]
            table[x expr={100 - \thisrow{test_ratio}}, y=cTF_avg]{\sctafdetopone};
            \addlegendentryexpanded{cTF};
    
            \addplot[teal,mark=square, mark size=4pt]
            plot [error bars/.cd, y dir=both, y explicit]
            table[x expr={100 - \thisrow{test_ratio}}, y=MoD_avg]{\sctafdetopone};
            \addlegendentryexpanded{MoD};
    
            \addplot[teal,mark=|, mark size=4pt]
            plot [error bars/.cd, y dir=both, y explicit]
            table[x expr={100 - \thisrow{test_ratio}}, y=cMoD_avg]{\sctafdetopone};
            \addlegendentryexpanded{cMoD};
            
            \end{axis}
        \end{tikzpicture}
    }

    \resizebox{.49\columnwidth}{!}{
    \begin{tikzpicture}
        \begin{axis}[
            xlabel={Train set Ratio (\%)},
            ylabel={Top-$1$ ADE (m)},
            label style={font=\Large},
            tick label style={font=\Large},
            legend columns=2,
            grid=major,
            ymax=1.15,
            xtick=data,
            legend style={nodes={scale=1.7, transform shape}}, 
        ]
        
        \addplot[magenta, mark=diamond, mark size=4pt]
        plot [error bars/.cd, y dir=both, y explicit]
        table[x expr={100 - \thisrow{test_ratio}}, y=LSTM_avg]{\sctbadetopone};
        \addlegendentryexpanded{RED};

        \addplot[magenta, mark=asterisk, mark size=4pt]
        plot [error bars/.cd, y dir=both, y explicit]
        table[x expr={100 - \thisrow{test_ratio}}, y=cLSTM_avg]{\sctbadetopone};
        \addlegendentryexpanded{cRED};

        \addplot[black, mark=oplus, mark size=4pt]
        plot [error bars/.cd, y dir=both, y explicit]
        table[x expr={100 - \thisrow{test_ratio}}, y=TF_avg]{\sctbadetopone};
        \addlegendentryexpanded{TF};

        \addplot[black, mark=triangle, mark size=4pt]
        plot [error bars/.cd, y dir=both, y explicit]
        table[x expr={100 - \thisrow{test_ratio}}, y=cTF_avg]{\sctbadetopone};
        \addlegendentryexpanded{cTF};

        \addplot[teal,mark=square, mark size=4pt]
        plot [error bars/.cd, y dir=both, y explicit]
        table[x expr={100 - \thisrow{test_ratio}}, y=MoD_avg]{\sctbadetopone};
        \addlegendentryexpanded{MoD};

        \addplot[teal,mark=|, mark size=4pt]
        plot [error bars/.cd, y dir=both, y explicit]
        table[x expr={100 - \thisrow{test_ratio}}, y=cMoD_avg]{\sctbadetopone};
        \addlegendentryexpanded{cMoD};
        
        \end{axis}
    \end{tikzpicture}
    }
    \resizebox{.49\columnwidth}{!}{
        \begin{tikzpicture}
            \begin{axis}[
                xlabel={Train set Ratio (\%)},
                ylabel={Top-$1$ FDE (m)},
                label style={font=\Large},
                tick label style={font=\Large},
                legend columns=2,
                grid=major,
                ymax=2.5,
                xtick=data,
                legend style={nodes={scale=1.7, transform shape}},
            ]
            
            \addplot[magenta, mark=diamond, mark size=4pt]
            plot [error bars/.cd, y dir=both, y explicit]
            table[x expr={100 - \thisrow{test_ratio}}, y=LSTM_avg]{\sctbfdetopone};
            \addlegendentryexpanded{RED};

            \addplot[magenta, mark=asterisk, mark size=4pt]
            plot [error bars/.cd, y dir=both, y explicit]
            table[x expr={100 - \thisrow{test_ratio}}, y=cLSTM_avg]{\sctbfdetopone};
            \addlegendentryexpanded{cRED};
    
            \addplot[black, mark=oplus, mark size=4pt]
            plot [error bars/.cd, y dir=both, y explicit]
            table[x expr={100 - \thisrow{test_ratio}}, y=TF_avg]{\sctbfdetopone};
            \addlegendentryexpanded{TF};

            \addplot[black, mark=triangle, mark size=4pt]
            plot [error bars/.cd, y dir=both, y explicit]
            table[x expr={100 - \thisrow{test_ratio}}, y=cTF_avg]{\sctbfdetopone};
            \addlegendentryexpanded{cTF};
    
            \addplot[teal,mark=square, mark size=4pt]
            plot [error bars/.cd, y dir=both, y explicit]
            table[x expr={100 - \thisrow{test_ratio}}, y=MoD_avg]{\sctbfdetopone};
            \addlegendentryexpanded{MoD};
    
            \addplot[teal,mark=|, mark size=4pt]
            plot [error bars/.cd, y dir=both, y explicit]
            table[x expr={100 - \thisrow{test_ratio}}, y=cMoD_avg]{\sctbfdetopone};
            \addlegendentryexpanded{cMoD};
            
            \end{axis}
        \end{tikzpicture}
    }
    \caption{
    Top-1 ADE/FDE scores in THÖR-MAGNI Scenario 3A (\textbf{top row}) and 3B (\textbf{bottom row}). In this class-balanced setting, deep learning methods surpass MoD approaches. However, MoD methods (MoD and cMoD) maintain stability even with reduced training data.}
    \label{fig:det_deep_learning_sc3ab}
    \vspace*{-4mm}
\end{figure}
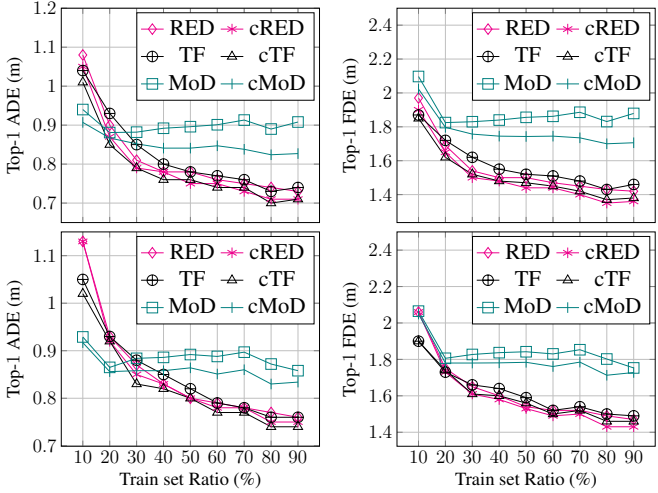

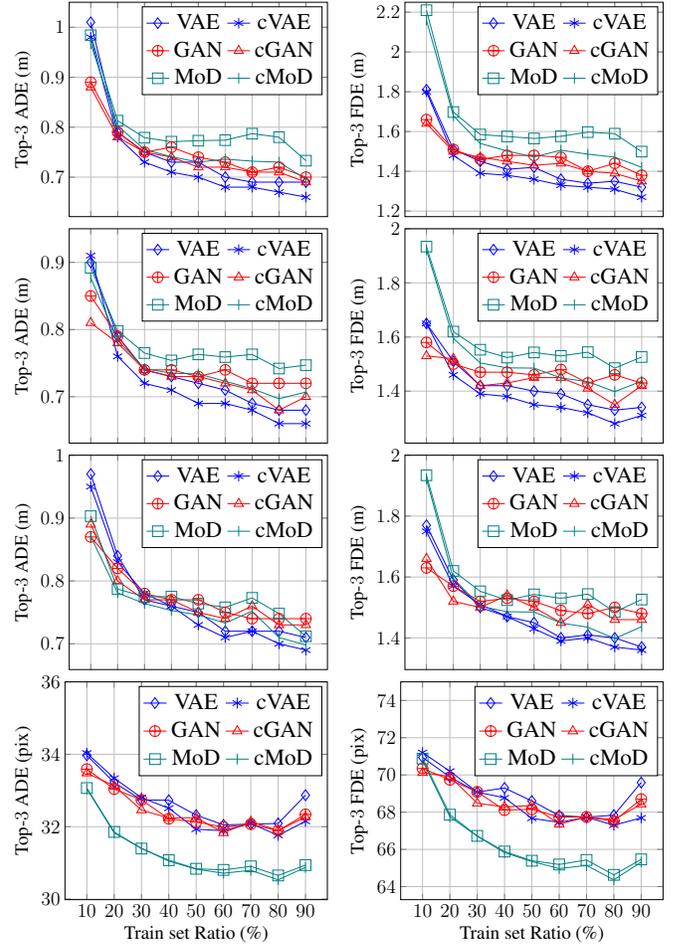
\begin{figure}[t]
    \centering
    \resizebox{.49\columnwidth}{!}{
    \begin{tikzpicture}
        \begin{axis}[
            ylabel={Top-$3$ ADE (m)},
            label style={font=\Large},
            tick label style={font=\Large},
            legend columns=2,
            grid=major,
            ymax=1.05,
            xtick=data,
            xticklabels={},
            legend style={nodes={scale=1.7, transform shape}}, 
        ]
        
        \addplot[blue, mark=diamond, mark size=4pt]
        plot [error bars/.cd, y dir=both, y explicit]
        table[x expr={100 - \thisrow{test_ratio}}, y=VAE_avg]{\sctwoadetopthree};
        \addlegendentryexpanded{VAE};

        \addplot[blue, mark=asterisk, mark size=4pt]
        plot [error bars/.cd, y dir=both, y explicit] 
        table[x expr={100 - \thisrow{test_ratio}}, y=cVAE_avg]{\sctwoadetopthree};
        \addlegendentryexpanded{cVAE};

        \addplot[red, mark=oplus, mark size=4pt]
        plot [error bars/.cd, y dir=both, y explicit] 
        table[x expr={100 - \thisrow{test_ratio}}, y=GAN_avg]{\sctwoadetopthree};
        \addlegendentryexpanded{GAN};

        \addplot[red, mark=triangle, mark size=4pt]
        plot [error bars/.cd, y dir=both, y explicit] 
        table[x expr={100 - \thisrow{test_ratio}}, y=cGAN_avg]{\sctwoadetopthree};
        \addlegendentryexpanded{cGAN};

        \addplot[teal,mark=square, mark size=4pt]
        plot [error bars/.cd, y dir=both, y explicit] 
        table[x expr={100 - \thisrow{test_ratio}}, y=MoD_avg]{\sctwoadetopthree};
        \addlegendentryexpanded{MoD};

        \addplot[teal,mark=|, mark size=4pt]
        plot [error bars/.cd, y dir=both, y explicit] 
        table[x expr={100 - \thisrow{test_ratio}}, y=cMoD_avg]{\sctwoadetopthree};
        \addlegendentryexpanded{cMoD};
        
        \end{axis}
    \end{tikzpicture}
    }
    \vspace*{-1mm}
    \resizebox{.49\columnwidth}{!}{
        \begin{tikzpicture}
            \begin{axis}[
                ylabel={Top-$3$ FDE (m)},
                label style={font=\Large},
                tick label style={font=\Large},
                legend columns=2,
                grid=major,
                ymax=2.25,
                xtick=data,
                xticklabels={},
                legend style={nodes={scale=1.7, transform shape}}, 
            ]
            
            \addplot[blue, mark=diamond, mark size=4pt]
            plot [error bars/.cd, y dir=both, y explicit]
            table[x expr={100 - \thisrow{test_ratio}}, y=VAE_avg]{\sctwofdetopthree};
            \addlegendentryexpanded{VAE};

            \addplot[blue, mark=asterisk, mark size=4pt]
            plot [error bars/.cd, y dir=both, y explicit] 
            table[x expr={100 - \thisrow{test_ratio}}, y=cVAE_avg]{\sctwofdetopthree};
            \addlegendentryexpanded{cVAE};
    
            \addplot[red, mark=oplus, mark size=4pt]
            plot [error bars/.cd, y dir=both, y explicit] 
            table[x expr={100 - \thisrow{test_ratio}}, y=GAN_avg]{\sctwofdetopthree};
            \addlegendentryexpanded{GAN};
    
            \addplot[red, mark=triangle, mark size=4pt]
            plot [error bars/.cd, y dir=both, y explicit] 
            table[x expr={100 - \thisrow{test_ratio}}, y=cGAN_avg]{\sctwofdetopthree};
            \addlegendentryexpanded{cGAN};

            \addplot[teal,mark=square, mark size=4pt]
            plot [error bars/.cd, y dir=both, y explicit] 
            table[x expr={100 - \thisrow{test_ratio}}, y=MoD_avg]{\sctwofdetopthree};
            \addlegendentryexpanded{MoD};
    
            \addplot[teal,mark=|, mark size=4pt]
            plot [error bars/.cd, y dir=both, y explicit] 
            table[x expr={100 - \thisrow{test_ratio}}, y=cMoD_avg]{\sctwofdetopthree};
            \addlegendentryexpanded{cMoD};
            
            \end{axis}
        \end{tikzpicture}
    }
    \vspace*{-1mm}
    \resizebox{.49\columnwidth}{!}{
    \begin{tikzpicture}
        \begin{axis}[
            ylabel={Top-$3$ ADE (m)},
            label style={font=\Large},
            tick label style={font=\Large},
            legend columns=2,
            grid=major,
            ymax=0.95,
            xtick=data,
            xticklabels={},
            legend style={nodes={scale=1.7, transform shape}}, 
        ]
        
        \addplot[blue, mark=diamond, mark size=4pt]
        plot [error bars/.cd, y dir=both, y explicit]
        table[x expr={100 - \thisrow{test_ratio}}, y=VAE_avg]{\sctaadetopthree};
        \addlegendentryexpanded{VAE};

        \addplot[blue, mark=asterisk, mark size=4pt]
        plot [error bars/.cd, y dir=both, y explicit] 
        table[x expr={100 - \thisrow{test_ratio}}, y=cVAE_avg]{\sctaadetopthree};
        \addlegendentryexpanded{cVAE};

        \addplot[red, mark=oplus, mark size=4pt]
        plot [error bars/.cd, y dir=both, y explicit] 
        table[x expr={100 - \thisrow{test_ratio}}, y=GAN_avg]{\sctaadetopthree};
        \addlegendentryexpanded{GAN};

        \addplot[red, mark=triangle, mark size=4pt]
        plot [error bars/.cd, y dir=both, y explicit] 
        table[x expr={100 - \thisrow{test_ratio}}, y=cGAN_avg]{\sctaadetopthree};
        \addlegendentryexpanded{cGAN};

        \addplot[teal,mark=square, mark size=4pt]
        plot [error bars/.cd, y dir=both, y explicit] 
        table[x expr={100 - \thisrow{test_ratio}}, y=MoD_avg]{\sctaadetopthree};
        \addlegendentryexpanded{MoD};

        \addplot[teal,mark=|, mark size=4pt]
        plot [error bars/.cd, y dir=both, y explicit] 
        table[x expr={100 - \thisrow{test_ratio}}, y=cMoD_avg]{\sctaadetopthree};
        \addlegendentryexpanded{cMoD};
        
        \end{axis}
    \end{tikzpicture}
    }
    \resizebox{.49\columnwidth}{!}{
        \begin{tikzpicture}
            \begin{axis}[
                ylabel={Top-$3$ FDE (m)},
                label style={font=\Large},
                tick label style={font=\Large},
                legend columns=2,
                grid=major,
                ymax=2.00,
                xtick=data,
                xticklabels={},
                legend style={nodes={scale=1.7, transform shape}}, 
            ]
            
            \addplot[blue, mark=diamond, mark size=4pt]
            plot [error bars/.cd, y dir=both, y explicit]
            table[x expr={100 - \thisrow{test_ratio}}, y=VAE_avg]{\sctafdetopthree};
            \addlegendentryexpanded{VAE};

            \addplot[blue, mark=asterisk, mark size=4pt]
            plot [error bars/.cd, y dir=both, y explicit] 
            table[x expr={100 - \thisrow{test_ratio}}, y=cVAE_avg]{\sctafdetopthree};
            \addlegendentryexpanded{cVAE};
    
            \addplot[red, mark=oplus, mark size=4pt]
            plot [error bars/.cd, y dir=both, y explicit] 
            table[x expr={100 - \thisrow{test_ratio}}, y=GAN_avg]{\sctafdetopthree};
            \addlegendentryexpanded{GAN};
    
            \addplot[red, mark=triangle, mark size=4pt]
            plot [error bars/.cd, y dir=both, y explicit] 
            table[x expr={100 - \thisrow{test_ratio}}, y=cGAN_avg]{\sctafdetopthree};
            \addlegendentryexpanded{cGAN};

            \addplot[teal,mark=square, mark size=4pt]
            plot [error bars/.cd, y dir=both, y explicit] 
            table[x expr={100 - \thisrow{test_ratio}}, y=MoD_avg]{\sctafdetopthree};
            \addlegendentryexpanded{MoD};
    
            \addplot[teal,mark=|, mark size=4pt]
            plot [error bars/.cd, y dir=both, y explicit] 
            table[x expr={100 - \thisrow{test_ratio}}, y=cMoD_avg]{\sctafdetopthree};
            \addlegendentryexpanded{cMoD};
            
            \end{axis}
        \end{tikzpicture}
    }
    \vspace*{-1mm}
    \resizebox{.49\columnwidth}{!}{
    \begin{tikzpicture}
        \begin{axis}[
            ylabel={Top-$3$ ADE (m)},
            label style={font=\Large},
            tick label style={font=\Large},
            legend columns=2,
            grid=major,
            ymax=1.0,
            xtick=data,
            xticklabels={},
            legend style={nodes={scale=1.7, transform shape}}, 
        ]
        
        \addplot[blue, mark=diamond, mark size=4pt]
        plot [error bars/.cd, y dir=both, y explicit]
        table[x expr={100 - \thisrow{test_ratio}}, y=VAE_avg]{\sctbadetopthree};
        \addlegendentryexpanded{VAE};

        \addplot[blue, mark=asterisk, mark size=4pt]
        plot [error bars/.cd, y dir=both, y explicit] 
        table[x expr={100 - \thisrow{test_ratio}}, y=cVAE_avg]{\sctbadetopthree};
        \addlegendentryexpanded{cVAE};

        \addplot[red, mark=oplus, mark size=4pt]
        plot [error bars/.cd, y dir=both, y explicit] 
        table[x expr={100 - \thisrow{test_ratio}}, y=GAN_avg]{\sctbadetopthree};
        \addlegendentryexpanded{GAN};

        \addplot[red, mark=triangle, mark size=4pt]
        plot [error bars/.cd, y dir=both, y explicit] 
        table[x expr={100 - \thisrow{test_ratio}}, y=cGAN_avg]{\sctbadetopthree};
        \addlegendentryexpanded{cGAN};

        \addplot[teal,mark=square, mark size=4pt]
        plot [error bars/.cd, y dir=both, y explicit] 
        table[x expr={100 - \thisrow{test_ratio}}, y=MoD_avg]{\sctbadetopthree};
        \addlegendentryexpanded{MoD};

        \addplot[teal,mark=|, mark size=4pt]
        plot [error bars/.cd, y dir=both, y explicit] 
        table[x expr={100 - \thisrow{test_ratio}}, y=cMoD_avg]{\sctbadetopthree};
        \addlegendentryexpanded{cMoD};

        \end{axis}
    \end{tikzpicture}
    }
    \resizebox{.49\columnwidth}{!}{
        \begin{tikzpicture}
            \begin{axis}[
                ylabel={Top-$3$ FDE (m)},
                label style={font=\Large},
                tick label style={font=\Large},
                legend columns=2,
                grid=major,
                ymax=2.00,
                xtick=data,
                xticklabels={},
                legend style={nodes={scale=1.7, transform shape}}, 
            ]
            
            \addplot[blue, mark=diamond, mark size=4pt]
            plot [error bars/.cd, y dir=both, y explicit]
            table[x expr={100 - \thisrow{test_ratio}}, y=VAE_avg]{\sctbfdetopthree};
            \addlegendentryexpanded{VAE};

            \addplot[blue, mark=asterisk, mark size=4pt]
            plot [error bars/.cd, y dir=both, y explicit] 
            table[x expr={100 - \thisrow{test_ratio}}, y=cVAE_avg]{\sctbfdetopthree};
            \addlegendentryexpanded{cVAE};
    
            \addplot[red, mark=oplus, mark size=4pt]
            plot [error bars/.cd, y dir=both, y explicit] 
            table[x expr={100 - \thisrow{test_ratio}}, y=GAN_avg]{\sctbfdetopthree};
            \addlegendentryexpanded{GAN};
    
            \addplot[red, mark=triangle, mark size=4pt]
            plot [error bars/.cd, y dir=both, y explicit] 
            table[x expr={100 - \thisrow{test_ratio}}, y=cGAN_avg]{\sctbfdetopthree};
            \addlegendentryexpanded{cGAN};

            \addplot[teal,mark=square, mark size=4pt]
            plot [error bars/.cd, y dir=both, y explicit] 
            table[x expr={100 - \thisrow{test_ratio}}, y=MoD_avg]{\sctafdetopthree};
            \addlegendentryexpanded{MoD};
    
            \addplot[teal,mark=|, mark size=4pt]
            plot [error bars/.cd, y dir=both, y explicit] 
            table[x expr={100 - \thisrow{test_ratio}}, y=cMoD_avg]{\sctafdetopthree};
            \addlegendentryexpanded{cMoD};

            \end{axis}
        \end{tikzpicture}

    }
    \resizebox{.49\columnwidth}{!}{
    \begin{tikzpicture}
        \begin{axis}[
            xlabel={Train set Ratio (\%)},
            ylabel={Top-$3$ ADE (pix)},
            label style={font=\Large},
            tick label style={font=\Large},
            legend columns=2,
            grid=major,
            ymax=36,
            xtick=data,
            legend style={nodes={scale=1.7, transform shape}}, 
        ]
        
        \addplot[blue, mark=diamond, mark size=4pt]
        plot [error bars/.cd, y dir=both, y explicit]
        table[x expr={100 - \thisrow{test_ratio}}, y=VAE_avg]{\sddadetopthree};
        \addlegendentryexpanded{VAE};

        \addplot[blue, mark=asterisk, mark size=4pt]
        plot [error bars/.cd, y dir=both, y explicit] 
        table[x expr={100 - \thisrow{test_ratio}}, y=cVAE_avg]{\sddadetopthree};
        \addlegendentryexpanded{cVAE};

        \addplot[red, mark=oplus, mark size=4pt]
        plot [error bars/.cd, y dir=both, y explicit] 
        table[x expr={100 - \thisrow{test_ratio}}, y=GAN_avg]{\sddadetopthree};
        \addlegendentryexpanded{GAN};

        \addplot[red, mark=triangle, mark size=4pt]
        plot [error bars/.cd, y dir=both, y explicit] 
        table[x expr={100 - \thisrow{test_ratio}}, y=cGAN_avg]{\sddadetopthree};
        \addlegendentryexpanded{cGAN};

        \addplot[teal,mark=square, mark size=4pt]
        plot [error bars/.cd, y dir=both, y explicit] 
        table[x expr={100 - \thisrow{test_ratio}}, y=MoD_avg]{\sddadetopthree};
        \addlegendentryexpanded{MoD};

        \addplot[teal,mark=|, mark size=4pt]
        plot [error bars/.cd, y dir=both, y explicit] 
        table[x expr={100 - \thisrow{test_ratio}}, y=cMoD_avg]{\sddadetopthree};
        \addlegendentryexpanded{cMoD};

        \end{axis}
    \end{tikzpicture}
    }
    \resizebox{.49\columnwidth}{!}{
        \begin{tikzpicture}
            \begin{axis}[
                xlabel={Train set Ratio (\%)},
                ylabel={Top-$3$ FDE (pix)},
                label style={font=\Large},
                tick label style={font=\Large},
                legend columns=2,
                grid=major,
                ymax=75,
                xtick=data,
                legend style={nodes={scale=1.7, transform shape}}, 
            ]
            
            \addplot[blue, mark=diamond, mark size=4pt]
            plot [error bars/.cd, y dir=both, y explicit]
            table[x expr={100 - \thisrow{test_ratio}}, y=VAE_avg]{\sddfdetopthree};
            \addlegendentryexpanded{VAE};

            \addplot[blue, mark=asterisk, mark size=4pt]
            plot [error bars/.cd, y dir=both, y explicit] 
            table[x expr={100 - \thisrow{test_ratio}}, y=cVAE_avg]{\sddfdetopthree};
            \addlegendentryexpanded{cVAE};

            \addplot[red, mark=oplus, mark size=4pt]
            plot [error bars/.cd, y dir=both, y explicit] 
            table[x expr={100 - \thisrow{test_ratio}}, y=GAN_avg]{\sddfdetopthree};
            \addlegendentryexpanded{GAN};
    
            \addplot[red, mark=triangle, mark size=4pt]
            plot [error bars/.cd, y dir=both, y explicit] 
            table[x expr={100 - \thisrow{test_ratio}}, y=cGAN_avg]{\sddfdetopthree};
            \addlegendentryexpanded{cGAN};

            \addplot[teal,mark=square, mark size=4pt]
            plot [error bars/.cd, y dir=both, y explicit] 
            table[x expr={100 - \thisrow{test_ratio}}, y=MoD_avg]{\sddfdetopthree};
            \addlegendentryexpanded{MoD};
    
            \addplot[teal,mark=|, mark size=4pt]
            plot [error bars/.cd, y dir=both, y explicit] 
            table[x expr={100 - \thisrow{test_ratio}}, y=cMoD_avg]{\sddfdetopthree};
            \addlegendentryexpanded{cMoD};
            
            \end{axis}
        \end{tikzpicture}
    }
    \caption{
    Top-3 ADE/FDE scores across TH\"{O}R-MAGNI Scenarios 2, 3A, 3B (\textbf{top to third rows}), and SDD (\textbf{bottom row}). In the class-balanced TH\"{O}R-MAGNI, deep generative methods excel over MoD. In the imbalanced SDD, MoD methods outperform deep generative methods across all data regimes.}
    \label{fig:gen_deep_learning_ade_sc23absdd}
    \vspace{-3mm}
\end{figure}

To assess how training data volume affects model performance, we conducted a data efficiency analysis aimed at identifying optimal models for various data settings.
\cref{fig:det_deep_learning_sc3ab} shows the performance of single-output methods (RED, TF, and MoD, along with their conditioned variants) in TH\"{O}R-MAGNI Scenarios 3A and 3B. cMoD outperforms deep learning methods in Top-$1$ ADE in low data regimes, where 10\% of data is available during training.
Moreover, performance for deep learning methods declines with less training data, whereas MoD approaches (MoD and cMoD) are more stable across different data regimes.
The MoD model we employ, CLiFF-map, efficiently captures major human motion patterns with limited training data. Beyond a 30\% training data increase, improvements in CLiFF-map are less notable, especially compared to the training set expansion from 10\% to 20\%. Once major motion patterns are captured, the representations stabilize, and unlike deep learning methods, MoD approaches do not show significant performance improvements. This stability highlights the MoD approach's advantage in scenarios where extensive data collection is impractical.
\cref{fig:gen_deep_learning_ade_sc23absdd} presents the performance of multiple-output methods (VAE, GAN, and MoD, along with their respective conditioned variants) on both datasets. In TH\"{O}R-MAGNI, deep generative methods prove more effective in generating one out of $K$ trajectories compared to MoD-based methods. Conversely, in the imbalanced dataset SDD, MoD-based methods consistently outperform deep generative methods across all train set ratios. These results underscore the preference for MoD-based methods for multiple outputs in imbalanced datasets.

\subsection{Qualitative Results}

\begin{figure}[!t]
\centering
\includegraphics[width=.48\linewidth]{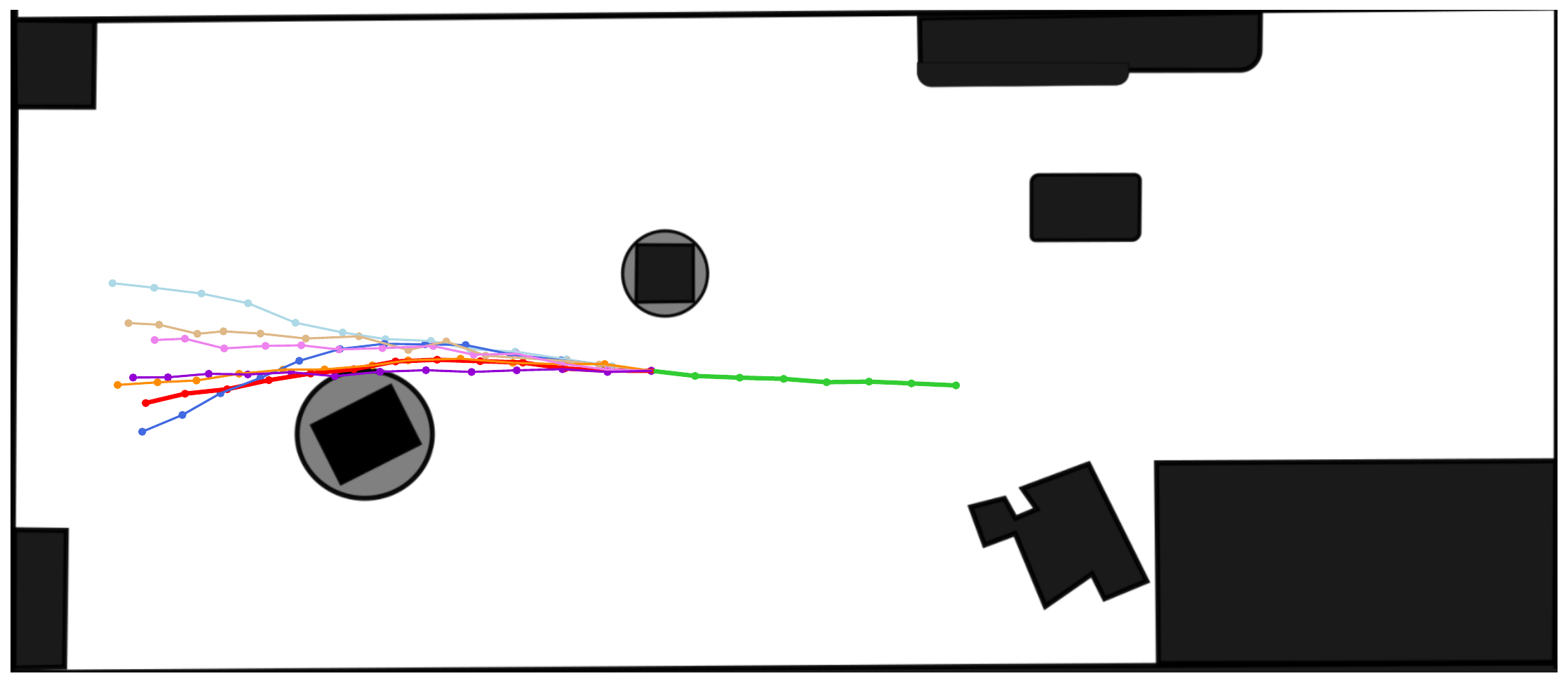}
\includegraphics[width=.48\linewidth]{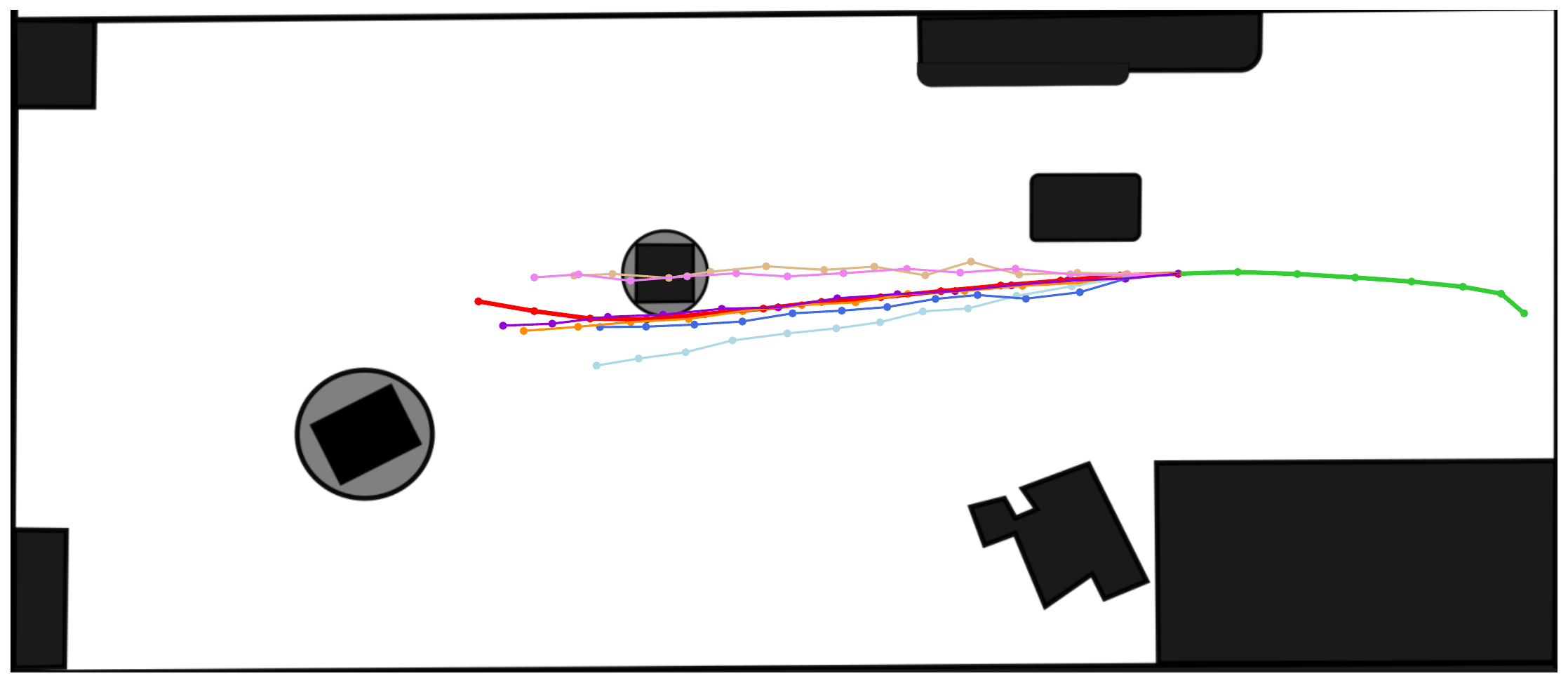}
\includegraphics[width=.48\linewidth]{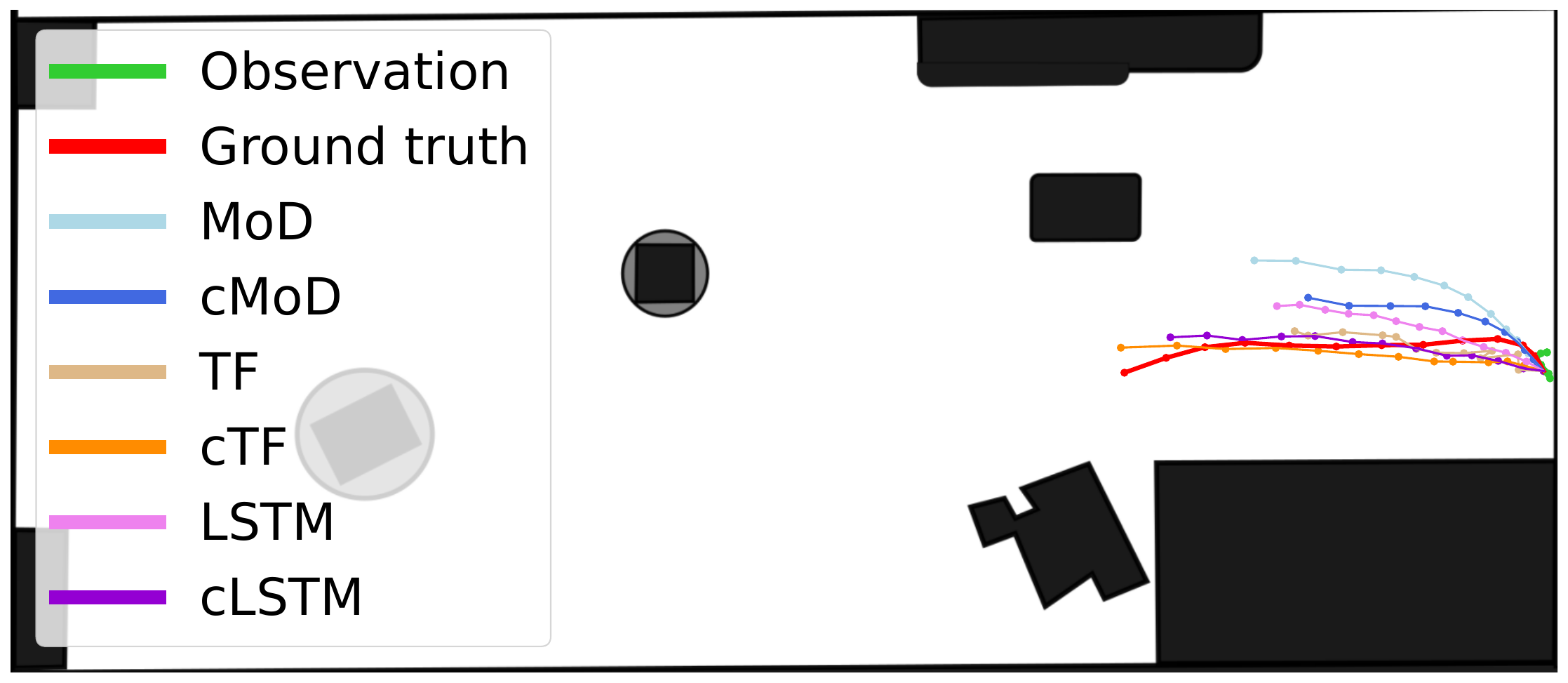}
\includegraphics[width=.48\linewidth]{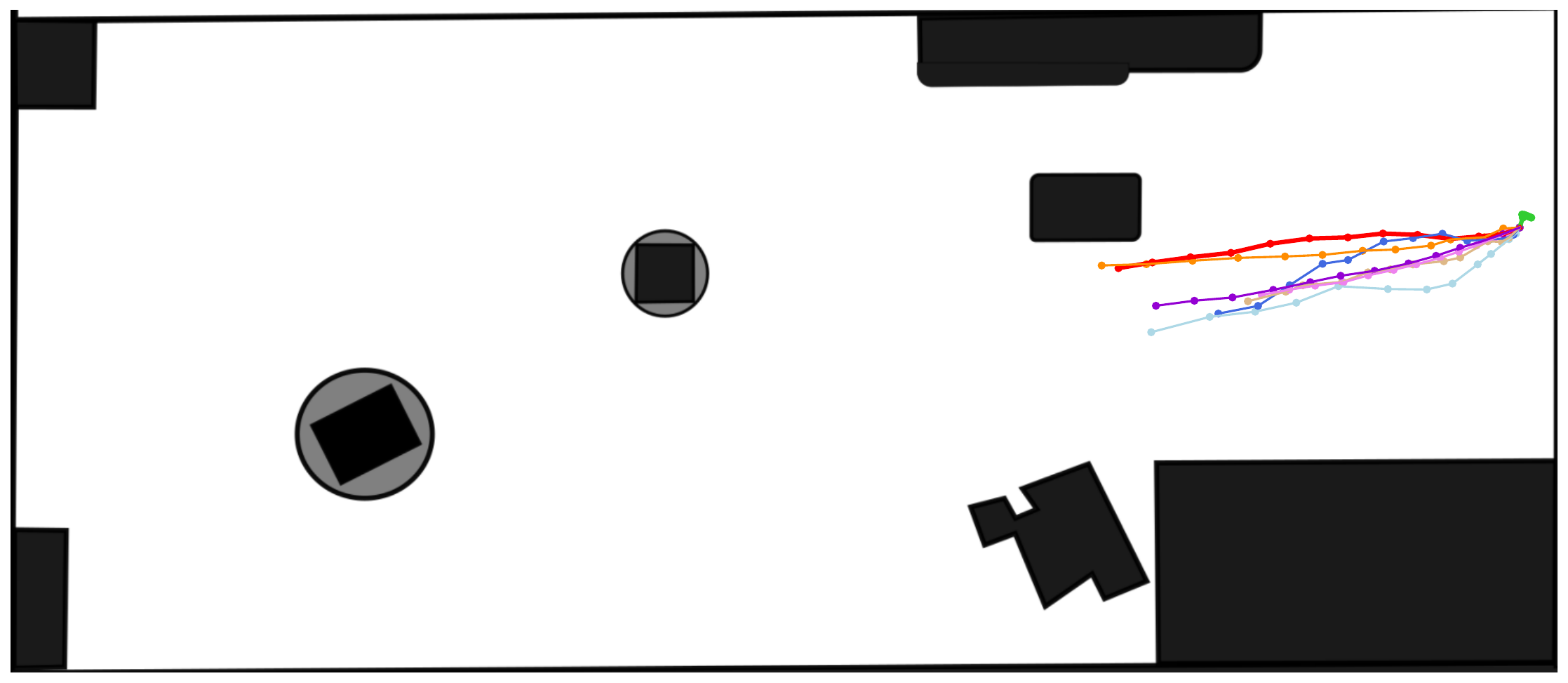}
\caption{Prediction examples of {\em Carrier--Box} (\textbf{top left}), {\em Carrier--Bucket} (\textbf{top right}), {\em Visitors--Alone} (\textbf{bottom left}) and {\em Carrier--Large Object} (\textbf{bottom right}) in TH\"{O}R-MAGNI with \SI{4.8}{\second} prediction horizon.}
\label{fig:magni_quality_res}
\vspace*{-4mm}
\end{figure}

We provide qualitative Top-$1$ trajectory prediction comparisons for each multiple-output approach in \cref{fig:sdd_quality_res} and for each single-output method in \cref{fig:magni_quality_res} for the SDD and TH\"{O}R-MAGNI datasets, respectively. For both datasets, conditioned methods are more accurate than their unconditional counterparts. On the SDD dataset, which is characterized by imbalanced classes, cMoD is the most effective compared to deep learning methods. On the TH\"{O}R-MAGNI dataset, we observed that conditioned deep learning methods outperform both unconditional deep learning methods and the MoD approaches, which is consistent with the quantitative results.

\subsection{Limitations}
In this work we analyze the effect of the dataset-imposed classes on motion prediction accuracy and compare deep learning with pattern-based methods across various data settings. However, our methods do not explicitly consider agent interactions, due to the own complexity of evaluating and comparing the interaction models \cite{rudenko2022atlas}. We aim to address this challenge in the future work.


\section{CONCLUSIONS \& FUTURE WORK} \label{section-conclusion}

The challenge of making accurate trajectory predictions in dynamic environments is further complicated when facing heterogeneous agents with diverse dynamics and distinct motion patterns. Considering the classes of agents can help lowering uncertainty in motion forecasts, an issue that arises when attempting to generalize across different classes. In this paper, we analyze how prior art in deep learning-based and pattern-based prediction can be adapted to consider class labels, concluding that class-conditioned methods generally outperform their unconditioned counterparts. 
The choice of a specific method, on the other hand, strongly depends on the available training data and the intended downstream application: in new environments with limited data, or where some classes are underrepresented and require multimodal predictions (sometimes critically so, e.g., vulnerable road users in automated driving), pattern-based methods may have an edge over the deep learning models. 
In future work, we plan to explore unsupervised trajectory and dynamics clustering to create more natural and informative class definitions. This approach aims to address the limitations of dataset-imposed classes (i.e., unstructured motion patterns within a class) and improve model performance in handling class imbalances.










\printbibliography
\end{document}